%% file: main_nature_MI.tex
\documentclass[letterpaper]{nature}

\usepackage{color}
\usepackage{url}
\usepackage{verbatim}
\usepackage{multirow}
\usepackage{xspace}
\usepackage{graphicx}
\usepackage{amsmath} 

\usepackage{amsthm}

\theoremstyle{definition}
\newtheorem{definition}{Definition}

\usepackage{thm-restate}

\usepackage{todonotes}
\usepackage{bbm}
\usepackage{amssymb}
\usepackage{booktabs} 
\usepackage{times} 
\usepackage{lscape}
\usepackage[left=1in,right=1in,bottom=1.1in,top=1in]{geometry}
\usepackage{enumitem}
\usepackage[normalem]{ulem}
\usepackage{xcolor}
\usepackage{colortbl}
\usepackage{array}
\usepackage[innercaption]{sidecap}
\usepackage{lineno}
\usepackage{stmaryrd}
\usepackage{xr-hyper}

\makeatletter
\newcommand*{\addFileDependency}[1]{
  \typeout{(#1)}
  \@addtofilelist{#1}
  \IfFileExists{#1}{}{\typeout{No file #1.}}
}
\makeatother

\newcommand*{\myexternaldocument}[1]{%
    \externaldocument{#1}%
    \addFileDependency{#1.tex}%
    \addFileDependency{#1.aux}%
}
\myexternaldocument{../SI/SI-main}

\usepackage[colorlinks,citecolor=blue,urlcolor=magenta]{hyperref}
\usepackage[margin=-1cm, labelfont=bf, font=small]{caption}

\newcommand{\hide}[1]{}
\newcommand{\cut}[1]{}

\newcommand{\xhdr}[1]{\vspace{1em}\noindent{{\bf #1}}}

\newcommand{\eg}{\emph{e.g.}\xspace}
\newcommand{\ie}{\emph{i.e.}\xspace}

\DeclareMathOperator*{\argmax}{argmax}
\let\oldnl\nl
\newcommand{\nonl}{\renewcommand{\nl}{\let\nl\oldnl}}  
\usepackage[normalem]{ulem}
\usepackage{comment}
\usepackage{bm}
\graphicspath{{./FIG/}}

\usepackage{amsmath}
\usepackage{amssymb}
\usepackage{multirow}
\usepackage{pdflscape}
\usepackage{listofitems}

\newcommand{\PT}{\mathrm{PT}}
\newcommand{\FT}{\mathrm{FT}}

\newcommand{\std}[1]{{\scriptsize$\pm$#1}}

\newcommand\res[1]{%
  \readlist*\args{#1}%
  \ifnum\argslen=1\relax%
    \args[1]
  \else%
    \ifnum\argslen=2\relax%
      \args[1]\std{\args[2]}
    \else
      \textbf{\args[1]\std{\args[2]}}
    \fi
  \fi%
}

\newcommand\resB[1]{%
  \readlist*\args{#1}%
  \ifnum\argslen=1\relax%
    \textbf{\args[1]}
  \else%
      \textbf{\args[1]\std{\args[2]}}
  \fi%
}
\newcommand{\imp}{$\boldsymbol{\uparrow}$}
\newcommand{\sam}{$\sim$}

\newcommand{\R}{\mathbb{R}}

\renewcommand{\vec}{\boldsymbol}
\newcommand{\mat}{\boldsymbol}

\renewcommand{\(}{\left(}
\renewcommand{\)}{\right)}
\newcommand{\ip}[1]{\left\langle#1\right\rangle}

\newcommand{\lossLM}{\mathcal L_\mathrm{M}}
\newcommand{\lossGGML}{\mathcal L_\mathrm{SI}}

\newcommand{\lambdaGGML}{\lambda_\mathrm{SI}}

\usepackage{caption}
\newcommand{\fullMethodName}{Structure Inducing Pre-Training\xspace}
\newcommand{\methodAbbr}{SIPT\xspace}
\newcommand{\LMPTAbbr}{LM PT\xspace} 

\newcommand{\notzeroshot}{Nearest-neighbor\xspace}

\definecolor{Gray}{gray}{0.85}
\definecolor{LightCyan}{rgb}{0.88,1,1}

\newcolumntype{g}{>{\columncolor{Gray}}c}

\usepackage{dsfont}
\usepackage{floatrow}
\newfloatcommand{capbtabbox}{table}[][\FBwidth]
\usepackage[normalem]{ulem}

\newcommand{\codelink}{https://github.com/mmcdermott/structure_inducing_pre-training}

\title{\begin{center}\fullMethodName\end{center}}    

\author{\begin{center}
Matthew B. A. McDermott$^{1,2}$, Brendan Yap$^{1}$, Peter Szolovits$^{1}$, Marinka Zitnik$^{2,3,4,\ddag}$\\[1mm]
\normalsize{$^{1}$Computer Science and Artificial Intelligence Laboratory, Massachusetts Institute of Technology, Cambridge, MA, 02139, USA} \\
\normalsize{$^{2}$Department of Biomedical Informatics, Harvard Medical School, Boston, MA, 02115, USA}\\
\normalsize{$^{3}$Broad Institute of MIT and Harvard, Cambridge, MA 02142, USA} \\
\normalsize{$^{4}$Harvard Data Science Initiative, Cambridge, MA 02138, USA} \\[4mm]  
\normalsize{$\ddag$Corresponding author. Email: marinka@hms.harvard.edu}
\end{center}}

\begin{document}
\nocite{ELMO_peters_deep_2018} \nocite{GPTIII_brown2020language} \nocite{T5_JMLR:v21:20-074} \nocite{liu_roberta_2019} \nocite{GPTI_radford2018improving} \nocite{GPTII_radford2019language} \nocite{BART_lewis-etal-2020-bart} \nocite{UNSUP_XLNG_conneau-etal-2020-unsupervised} \nocite{ELECTRA_clark_electra_2019} \nocite{SPANBERT_joshi-etal-2020-spanbert} \nocite{UNI_LM_NEURIPS2019_c20bb2d9} \nocite{DAPT_gururangan-etal-2020-dont} \nocite{ERNIE1_sun_ernie_2019} \nocite{KNOWBERT_peters2019knowledge} \nocite{rao_evaluating_2019} \nocite{LUKEyamada2020luke} \nocite{T0PP_sanh2022multitask} \nocite{PRE_ENC_Xiong2020Pretrained} \nocite{MSA_pmlr-v139-rao21a} \nocite{COLAKE_sun-etal-2020-colake} \nocite{BERTMK_he-etal-2020-bert} \nocite{ERICA_qin2020erica} \nocite{JAKET_yu2020jaket} \nocite{CALM_zhou2021pretraining} \nocite{KEBIOLM_yuan-etal-2021-improving} \nocite{MGBERT_zhang2021mg} \nocite{CDLM_caciularu-etal-2021-cdlm-cross} \nocite{KGPLM_he2020kgplm} \nocite{KNN_PT_levine2022the} \nocite{LP-BERT_DBLP:journals/corr/abs-2201-04843} \nocite{MGBERT_behnamghader2021mg} \nocite{UD-PRLM_NEURIPS2021_473803f0} \nocite{rives_biological_2019} \nocite{alley_unified_2019} \nocite{devlin_bert_2019} \nocite{ERNIE_ALT_zhang_ernie_2019} \nocite{COKEBERT_su2021cokebert} \nocite{SPIDER_zhang-zhao-2021-structural} \nocite{kuncoro-etal-2020-syntactic} \nocite{lan_albert_2019} \nocite{SMEDBERT_zhang2021smedbert} \nocite{MT_DNN_liu_multi-task_2019} \nocite{GRAPH_PT_hu_strategies_2019} \nocite{SENTILARE_ke-etal-2020-sentilare} \nocite{PLUS_min_pre-training_2020} \nocite{mcdermott_comprehensive_2020} \nocite{ERNIE2_sun_ernie_2020} \nocite{ERNIE3_sun2021ernie} \nocite{Dict-BERT-yu-etal-2022-dict} \nocite{LINKBERT_yasunaga-etal-2022-linkbert} \nocite{STRUCTBERT_Wang2020StructBERT} \nocite{MARGE_lewis2020pre} \nocite{REALM_10.5555/3524938.3525306} \nocite{GraphCL_NEURIPS2020_3fe23034} \nocite{GCC_10.1145/3394486.3403168} \nocite{DECLUTR_giorgi-etal-2021-declutr} \nocite{CLEAR_wu2020clear} \nocite{JOAO_pmlr-v139-you21a} \nocite{COCO_LM_NEURIPS2021_c2c2a045} \nocite{INFOWORD_Kong2020A} \nocite{MICRO_GRAPH_zhang2020motif} \nocite{STS_CT_carlsson2021semantic} \nocite{CAPT_luo2020capt} \nocite{GEARNET_zhang2022protein} \nocite{INFOXLM_chi-etal-2021-infoxlm} \nocite{GLM_shen-etal-2020-exploiting} \nocite{KCL_fang2022molecular} \nocite{wang_kepler_2019} \nocite{CK_GNN_fang2021knowledge} \nocite{XLM_K_jiang2022xlm} \nocite{WEBFORMER_guo2022webformer}

\maketitle
{\spacing{1.4} 
\begin{abstract}
\input{000abstract}
\end{abstract}
}

\clearpage

\spacing{1.2}

\section*{Main}

\input{010intro}

\section*{Results}\label{sec:results}
\input{020results.tex}

\section*{Discussion}\label{sec:conclusion}
\input{030discussion}

\clearpage

\paragraph{Data availability.}

Our synthetic datasets and pointers to all real-world datasets used (which are all publicly available) are available here: \url{\codelink}.

\paragraph{Code availability.}

All code for this project is available at \url{\codelink}.

\paragraph{Acknowledgements.} 

MBAM was partly supported by a National Institutes of Health (NIH) grant LM013337 and a collaborative research agreement with IBM.
%
BY was supported by a Massachusetts Institute of Technology (MIT) Undergraduate Research Opportunity fund.
%
MZ gratefully acknowledges the support by the NSF under Nos. IIS-2030459 and IIS-2033384, US Air Force Contract No. FA8702-15-D-0001, and awards from Harvard Data Science Initiative, Amazon Research, Bayer Early Excellence in Science, AstraZeneca Research, and Roche Alliance with Distinguished Scientists. Any opinions, findings, conclusions or recommendations expressed in this material are those of the authors and do not necessarily reflect the views of the funders.

\paragraph{Authors contribution.} 

MBAM and BY collated datasets, wrote modelling code, and ran experiments. MBAM compiled final results and completed the review of existing pre-training studies. MBAM, PS, and MZ conceived of the study and shaped the framing of the work. PS and MZ offered insight and guidance throughout the project. MBAM and MZ wrote the final manuscript, and MBAM, BY, PS, and MZ contributed edits to manuscript drafts. 

\paragraph{Competing interests.} 

The authors declare no competing interests.

\clearpage

\input{040figures}
\input{041methodsfigures}

\clearpage

\input{042methodstables}

\clearpage 

\section*{Online Methods}\label{sec:suppl_methods}
\input{050online_methods}

\clearpage

\section*{References}

{
\spacing{0.85}
\bibliographystyle{naturemag}
\bibliography{refs}
}

\clearpage

\appendix
\input{060appendix}

\end{document}

%% file: 000abstract.tex
Language model pre-training and derived methods are incredibly impactful in machine learning.
However, there remains considerable uncertainty on exactly why pre-training helps improve performance for fine-tuning tasks. This is especially true when attempting to adapt language-model pre-training to domains outside of natural language.
Here, we analyze this problem by exploring how existing pre-training methods impose relational structure in their induced per-sample latent spaces---\ie, what constraints do pre-training methods impose on the distance or geometry between the pre-trained embeddings of two samples $\vec x_i$ and $\vec x_j$.
Through a comprehensive review of existing pre-training methods, we find that this question remains open. This is true despite theoretical analyses demonstrating the importance of understanding this form of induced structure.
Based on this review, we introduce a descriptive framework for pre-training that allows for a granular, comprehensive understanding of how relational structure can be induced. We present a theoretical analysis of this framework from first principles and establish a connection between the relational inductive bias of pre-training and fine-tuning performance. We also show how to use the framework to define new pre-training methods.
We build upon these findings with empirical studies on benchmarks spanning 3 data modalities and ten fine-tuning tasks. These experiments validate our theoretical analyses, inform the design of novel pre-training methods, and establish consistent improvements over a compelling suite of baseline methods.

%% file: 010intro.tex

The pre-training (PT)/fine-tuning (FT) learning paradigm (also known as transfer learning) has had tremendous impact on natural language processing (NLP) and related domains~\cite{devlin_bert_2019,deng2009imagenet,GPTIII_brown2020language}. 
In NLP or NLP-derived PT/FT, we are given a dataset $\mat X \in \mathcal X^{N_\PT}$ and attempt to pre-train an encoder $f_{\vec \theta}: \mathcal X \to \mathcal Z$ which maps our domain of interest $\mathcal X$ into a latent space $\mathcal Z$: $f_{\vec \theta}: \vec x_i \mapsto \vec z_i$. This encoder $f_{\vec \theta}$ is then transferred for use in various fine-tuning tasks (which are not known at pre-training time). 
We evaluate PT/FT systems via the transfer performance of $f_{\vec \theta}$ on said fine-tuning tasks. 

In this work, we are concerned primarily with the efficacy of PT/FT for downstream tasks that operate at a \emph{per-sample} level (\eg, in natural language processing, evaluating the sentiment of a whole restaurant review is a \emph{per-sample} task, in contrast to identifying a named entity token within a sentence which is an \emph{intra-sample}/\emph{per-token} task). One aspect of pre-training that drives such eventual fine-tuning performance is the induced geometry of the pre-trained, per-sample latent space $\mathcal Z$ (formally defined in the Methods section).
For example, it is well documented that the sentence embeddings produced by pre-trained language models in NLP can be non-smooth and anisotropic, which harms downstream task performance~\cite{BERT_FLOW_li-etal-2020-sentence}. In other domains, such as biomedical modalities, where per-sample tasks are even more prevalent than intra-sample tasks as compared to NLP, the importance of this geometry only increases. 
Despite this importance, research into mechanisms to induce explicit, deep structural constraints in $\mathcal Z$ is surprisingly limited. Many methods outright ignore the geometry of $\mathcal Z$ (\eg, by imposing no pre-training loss over the whole-sample embeddings during pre-training)~\cite{liu_roberta_2019,GPTI_radford2018improving,GPTI_radford2018improving,GPTIII_brown2020language} and other methods impose either only shallow structural constraints, such as through an auxiliary, per-sample, classification PT objective~\cite{devlin_bert_2019,lan_albert_2019,MT_DNN_liu_multi-task_2019}, or deeper structural constraints, but in an implicit manner, such as through data-augmentation~\cite{DECLUTR_giorgi-etal-2021-declutr,INFOWORD_Kong2020A} or noising-based contrastive losses~\cite{CLEAR_wu2020clear,COCO_LM_NEURIPS2021_c2c2a045}.
While such methods can be powerful and have been successful in many areas, we argue that the lack of a clear framework to design PT methods that impose structural constraints on $\mathcal Z$ that are simultaneously \emph{explicit} (similar to supervised classification losses) and \emph{deep} (similar to noising/augmentation-based contrastive losses) is a major weakness.

On the basis of this observation, we develop an analytical framework under which the PT objective is subdivided into two components: first, a language-model inspired imputation/denoising objective that leverages intra-sample relationships, and, second, a loss term explicitly driven to regularize the geometry of the per-sample latent space $\mathcal Z$ to reflect the connectivity patterns of a user-specified graph $G_\PT$. By relying on graphs to capture the structure we wish to induce in $\mathcal Z$, this PT framework allows us to specify PT methods that induce \emph{deep} structure in an \emph{explicit} manner, filling exactly the gap identified above. In addition, this paradigm can capture diverse relationships, such as those motivated by external knowledge (\eg,~\cite{zitnik_evolution_2019}), self-supervised constraints (\eg,~\cite{wang_review_2019,hu_open_2020}), or distances between samples in an alternate modality (\eg,~\cite{CK_GNN_fang2021knowledge}). Moreover, this PT framework is simultaneously specific enough to allow us to make theoretical guarantees about how different PT graphs impact FT performance, general enough to encompass a variety of existing PT methods, and expressive enough to motivate new PT methods that have not been previously studied. In addition to theoretical analysis, we demonstrate empirically that defining new methods according to our framework, using explicit forms of real-world structure, yields significant benefits over competitive PT baselines across 3 modalities and 10 FT tasks. 

Our work advances PT/FT research through three major contributions.
First, we show via a comprehensive review and detailed commentary that existing pre-training methods largely do not induce structural constraints over $\mathcal Z$ that are simultaneously \emph{deep} and \emph{explicit}.
Second, we establish a new framework for describing PT methods, which provides a vehicle to design new PT methods that explicitly induce deep structural constraints in $\mathcal Z$ in accordance with a user-specified PT graph $G_\PT$. We further support this framework with theoretical results quantifying how the graph's structure relates to FT task performance. Crucially, this formalization in our new PT paradigm offers insight into when PT does or does not add value over supervised learning alone.
Third, we show that structure-inducing PT methods through our framework perform at or above the level of existing PT baselines across three data modalities and 10 FT tasks.

%% file: 020results.tex
\subsection*{General Pre-Training Problem Formulation}\label{sec:problem-formulation}

Given a dataset $\mat X_\PT \in \mathcal{X}^{N_\PT}$, a PT method aims to learn an encoder $f_{\vec \theta} : \mathcal X \to \mathcal Z$ such that $f_{\vec \theta}$ can be transferred to FT tasks that are unknown at pre-training time.
While we can leverage additional information at PT time to inform the training of $f_{\vec \theta}$ (\eg, PT-specific labels $\mat Y_\PT$), the encoder $f_{\vec \theta}$ \emph{must take only samples from $\mathcal X$ as inputs so that it can be used for fine-tuning}. Pre-training methods typically solve this problem by training $f_{\vec \theta}$ to minimize a pre-training loss $\mathcal L_\PT$ over $\mat X_\PT$. 
For example, in BERT, $\mathcal X$ consists of free-text samples, $f_{\vec \theta}$ is a transformer model, and $\mathcal L_\PT$ consists of both a masked language modelling (MLM) per-token loss and the next-sentence-prediction (NSP) per-sample loss~\cite{devlin_bert_2019}. 

Note that our definition of pre-training ignores secondary applications of the pre-training objective itself; for example, autoregressive language models (\eg, GPT-3~\cite{GPTIII_brown2020language}) are often used for their generative use directly, and not as commonly used to acquire embeddings or in transfer learning. This is a perfectly valid use of pre-trained language models within NLP, but is often not as useful in other domains which lack NLP's generative properties, so we focus on the induced embeddings produced by pre-training methods instead. Note further that we are primarily interested in PT methods that either are or are derived from NLP PT methods. This domain is of particular interest because these methods (1) have been extremely successful within NLP~\cite{devlin_bert_2019,GPTIII_brown2020language,T0pp_sanh2021multitask}, (2) have motivated a large number of derived methods in non-language, biomedical modalities~\cite{rives_biological_2019,GRAPH_PT_hu_strategies_2019,mcdermott_comprehensive_2020,MSA_pmlr-v139-rao21a}, and (3) are not yet fully technically understood~\cite{saunshi_theoretical_2019,KNN_PT_levine2022the,BERT_FLOW_li-etal-2020-sentence}.

\subsection*{Defining Explicit and Deep Structural Constraints}\label{sec:pt-problem}
Central to our hypothesis is the claim that most NLP-derived PT methods today do not impose explicit, deep constraints on the (per-sample) latent space geometry of $\mathcal Z$. To justify this claim, we define ``explicit'' and ``deep'' structural constraints (Definitions~\ref{def:explicit-implicit}-\ref{def:deep-shallow}). 

\begin{definition}{Explicit vs. Implicit Structural Constraints:} \label{def:explicit-implicit}

A PT objective $\mathcal L_\PT$ imposes a structural constraint that is \emph{explicit} (vs. implicit) to the degree that it (as $f_{\vec \theta}$ approaches optimality) permits us to reason directly about the relationship (in particular, the distance) between any two samples $\vec z_i$ and $\vec z_j$ in the latent space $\mathcal Z$.
\end{definition}

\begin{definition}{Deep vs. Shallow Structural Constraints:}
\label{def:deep-shallow}

A PT objective $\mathcal L_\PT$ imposes a structural constraint that is \emph{deep} (vs. shallow) on the basis of how much information (\eg, how many dimensions) would be required to fully satisfy the constraint.
\end{definition}

For example, consider a classification PT loss according to labels $y_i \in \mathcal Y$ and a logit layer which maps $\vec z_i \mapsto \tilde{y_i}$. This method produces an \emph{explicit} structural constraint because near optimality, we can infer that the relative (cosine) distance between two samples $\vec z_i$ and $\vec z_j$ is small if and only if $y_i = y_j$. However, this constraint is also \emph{shallow}, because to fully satisfy this constraint, we need only embed each class $c \in \mathcal Y$ with a unique position $\vec p_c \in \mathcal Z$, then compress all samples $\vec z_i$ near their class prototype $\vec p_{y_i}$. This distance-based constraint can be accomplished in a very low dimensional space $\mathcal Z$ (\eg we can distribute each $\vec p_c$ uniformly about a 2D unit circle, then compress all $\vec z_i$ to appear at a very small cosine distance from their class prototypes), illustrating that this constraint is very shallow.

In contrast, consider a contrastive method that asserts that $\vec z_i = f_{\vec \theta}(\vec x_i)$ should be close to $\vec z_i' = f_{\vec \theta}(\Tilde{\vec x_i})$, under some noising/augmentation procedure $\vec x_i \mapsto \Tilde{\vec x_i}$, but simultaneously far from other samples $\vec z_j$. While this method constrains the latent space to be smooth with respect to the noising process, it offers only an \emph{implicit} constraint on $\mathcal Z$ as it is generally not possible to infer how the distance between distinct samples $\vec z_i$ and $\vec z_j$ is constrained. However, it imposes a \emph{deeper} constraint than does the classification objective because the implicit connections between samples induced by the noising procedure reflect relationships that can not necessarily be captured in a low-dimensional space (dependent on dataset size and density).

\subsection*{Existing Pre-training Methods do not use Deep, Explicit Constraints}
To show that existing methods largely do not provide means to impose structural constraints that are simultaneously deep and explicit, we survey over 90 existing PT methods on the basis of how their objective functions constrain the $\mathcal Z$ (Figure~\ref{fig:existing_pt_methods}, Appendix \ref{sec:existing_LMPT_details}). For full details on our review findings, see the Methods section.
Throughout all examined methods, we find that \emph{deep, explicit structural constraints are almost never employed}. Instead, most methods either (1) impose no per-sample PT objectives at all (\eg, text-generation models, which are often not used for embeddings at all but rather for prompting or generative applications~\cite{GPTI_radford2018improving,GPTII_radford2019language,GPTIII_brown2020language,liu_roberta_2019}), (2) use explicit, but shallow, supervised PT objectives (\eg, BERT's ``Next-sentence Prediction'' (NSP) objective, ALBERT's ``Sentence-order Prediction'' (SOP) objective, or various multi-task objectives~\cite{devlin_bert_2019,lan_albert_2019,MT_DNN_liu_multi-task_2019}), or (3) use implicit, but deep, un- or self-supervised contrastive PT objectives (\eg, contrastive sentence embedding losses~\cite{DECLUTR_giorgi-etal-2021-declutr, INFOWORD_Kong2020A,CLEAR_wu2020clear,AKEBERT_ribeiro2021combining,COCO_LM_NEURIPS2021_c2c2a045}).

Across all surveyed methods, we find that only four methods impose simultaneously explicit and deep constraints: KEPLER~\cite{wang_kepler_2019}, CK-GNN~\cite{CK_GNN_fang2021knowledge}, XLM-K~\cite{XLM_K_jiang2022xlm}, and WebFormer~\cite{WEBFORMER_guo2022webformer}. All four can be described as some form of per-sample graph alignment, in which an external, pre-training knowledge graph $G_\PT$ or connectivity algorithm is employed over a subset of pre-training samples, and the output embeddings of pairs of samples $\vec z_i = f_{\vec \theta}(\vec x_i)$ and $\vec z_j = f_{\vec \theta}(\vec x_j)$ are constrained to reflect their relationships in the pre-training graph. This form of constraint is explicit, as the graph $G_\PT$ contains explicit relationships that will be induced in the output latent space, but also deep, as the geometry of the graph $G_\PT$ can be arbitrarily complex.

However, all these methods have major limitations.
In KEPLER and XLM-K, the per-sample embeddings are only constrained to a restricted set of samples corresponding to entity descriptions from a knowledge graph. As such, there are no constraints implied on the general domain free-text samples in $\mathcal X$ alone~\cite{wang_kepler_2019,XLM_K_jiang2022xlm}. 
In CK-GNN, the graph connectivity is derived from a cluster-restricted 1-nearest-neighbor graph in an alternate modality's distance space, which may offer a limited higher-order structure, and unlike the NLP approaches, this method has no intra-sample (\eg per-token) pre-training task~\cite{CK_GNN_fang2021knowledge}.
Finally, in WebFormer, the graph used is inferred from the structure of the HyperText Markup Language (HTML) underlying web-pages, and relationships are only constrained at the per-sample level for limited structural relationships within the HTML. Further, WebFormer is a specialized model specifically for processing web content (text and HTML elements), so their approach can't be directly generalized to other domains~\cite{WEBFORMER_guo2022webformer}.
Moreover, these methods explore only the particular contexts of their individual models. They offer no general framework for how to realize this style of deep, explicit per-sample constraints in other contexts, nor do they explore any theory on how these constraints relate to performance for fine-tuning tasks~\cite{wang_kepler_2019,CK_GNN_fang2021knowledge,XLM_K_jiang2022xlm,WEBFORMER_guo2022webformer}.

Overall, our review of pre-training methods establishes unequivocally that pre-training methods capable of providing explicit, deep structural constraints are significantly under-explored. Across all the methods we reviewed, only four methods leverage constraints are explicit and deep, all of which have significant limitations, and there is no general consensus on how to constrain the $\mathcal Z$ explicitly and deeply. These findings motivate our new framework, which offers insight into how to realize deep, explicit structural constraints in pre-training models across diverse contexts and provides theoretical guidance on how structural constraints relate to fine-tuning performance.

\subsection*{New Pre-training Framework: Structure-Inducing Pre-training (SIPT)}
Our pre-training problem framework includes two small, but important, differences from the standard formulation (Figure~\ref{fig:framework_overview}).

First, we assume that we have as an additional input to the PT problem a graph $G_\PT = (V, E)$ where vertices denote pre-training samples within $\mat X_\PT$ (\eg, $\{\vec x_\PT | \vec x_\PT \in \mat X_\PT\} \subseteq V$) and edges represent user-specified relationships. Importantly, while we take the graph $G_\PT$ an input to the PT problem, \emph{we cannot use it as a direct input to $f_{\vec\theta}$}. Just like in traditional pre-training, $f_{\vec \theta}$ must take as input only samples from $\mathcal X$. \emph{This is because otherwise, we can not apply $f_{\vec \theta}$ to the same, general class of FT tasks over domain $\mathcal X$}. 

Second, we decompose the PT loss $\mathcal L_\PT$ into two components, weighted with hyperparameter $0 \le \lambdaGGML \le 1$:  
\[\mathcal L_\PT = (1 - \lambdaGGML) \lossLM + \lambdaGGML \lossGGML.\]
$\lossLM$ is a traditional, intra-sample objective (\eg, a language model), and $\lossGGML$ is a new, structure-inducing objective designed to regularize the per-sample latent space geometry in accordance with the relationships (edges) in $G_\PT$.
Under our framework, $\lossGGML$ is only allowable for $G_\PT$, $f_{\vec \theta}$, and $\mathcal Z$ if it permits some stable optima at which point a radius nearest-neighbor connectivity algorithm under some distance function in $\mathcal Z$ will recover $G_\PT$ (formal constraint is in the Methods section). Note that this constraint strikes a connection between our framework and the wealth of existing research focused on \emph{graph representation learning}~\cite{gao2018deep,cui2020adaptive,li2018community,li2021representation,GCNN_KIPF_WELLING_DBLP:conf/iclr/KipfW17,GRAPHSAGE_hamilton2017inductive}. These techniques do indeed offer valuable insights into how to sample minibatches over graph-structured data and devise losses for graph embeddings; however, many methods for actually modelling graph-structured data, including deep attributed graph embeddings and graph convolutional neural networks, should not be seen as replacements for our techniques here as they are typically not adaptable to contexts in which the graph is not known at inference time, and so \emph{they could not be used in our pre-training setting where $f_{\vec\theta}$ must take in only inputs from $\mathcal X$ directly}.

As the new loss term added $\lossGGML$ is explicitly designed to \emph{induce the structure of $G_\PT$ in $\mathcal Z$}, we call methods trained under our framework \emph{structure-inducing pre-training} (SIPT) methods. Many existing PT approaches can be re-realized as SIPT methods, including classification-based PT objectives like NSP or SOP, contrastive methods, or existing graph alignment methods (see Methods for full details).

\subsection*{Theoretical Analyses}\label{sec:theoretical_results}

Under our framework, one can link the structure of the PT graph $G_\PT$ to eventual FT task performance. In particular, as a SIPT embedder $f$ over graph $G_\PT$ approaches optimality under the loss $\lossGGML$, it produces an embedding space such that nearest-neighbor performance for any downstream task is lower bounded by the performance that could be obtained via a nearest neighbor algorithm over graph $G_\PT$ (Theorem~\ref{thm:nn_acc_to_LC}). This fact directly connects the geometry of the graph $G_\PT$ with the eventual fine-tuning performance of a SIPT embedder $f$. Furthermore, it demonstrates the advantage of employing an explicit constraint rather than an implicit one; by controlling the structure of $G_\PT$, users can directly choose to add different inductive biases to the PT process, in a manner which has a provable impact on the eventual suitability for downstream FT tasks.
\begin{restatable}{thm}{thmOne}
Let $\mat X_\PT$ be a PT dataset, $G_\PT$ be a PT graph, and let $f_{\vec \theta^*}$ be an encoder pre-trained under a PT objective permissible under our framing that realizes a $\lossGGML$ value no more than $\ell^*$. 
Then, under embedder $f$, the nearest-neighbor accuracy for a FT task $y$ converges as dataset size increases to at least the local consistency (Definition~\ref{def:LC}) of $y$ over $G_\PT$.
\label{thm:nn_acc_to_LC}
\end{restatable}

We also establish two important corollaries of Theorem~\ref{thm:nn_acc_to_LC} that further illustrate the importance of choosing graphs $G_\PT$ which impose \emph{deep} structural constraints (Corollaries~\ref{corollary:cliques_shallow}-\ref{corollary:manifolds_deep}).
\begin{restatable}{coro}{coroCliques}
    Let $\mat X_\PT \in \mathcal X^N$, be a PT dataset with corresponding labels $\vec y \in \mathcal Y_\PT^N$. Define $G_\PT = (\mat X_\PT, E)$ such that $(\vec x_i, \vec x_j) \in E$ if and only if $y_i = y_j$.
    
    Then, the local consistency for a given FT task $\vec y^{\text{(FT)}}$ over $G_\PT$ (and thus by Theorem~\ref{thm:nn_acc_to_LC}, the nearest-neighbor accuracy for any optimized SIPT embedder) is upper bounded by the probability that a sample $x_i$'s fine-tuning label $y^{\text{(FT)}}_i$ agrees with the majority class label for task $\vec y^{\text{(FT)}}$ over the clique consisting of all nodes with the same \emph{pre-training} label $y_i$ as $x_i$.
    \label{corollary:cliques_shallow}
\end{restatable}
\begin{restatable}{coro}{coroManifolds}
    Let $\mat X_\PT$ be a PT dataset that can be realized over a valid manifold $\mathcal M$. Assume $\mat X_\PT$ is sampled with full support over $\mathcal M$. Let $G_\PT (\mat X_\PT, E)$ be an $r$-nearest-neighbor graph over $\mathcal M$ (\eg, $(\vec x_i, \vec x_j) \in E$ if and only if the geodesic distance between the two points on $\mathcal M$ is less than $r$: $\mathcal D_{\mathcal M}(\vec x_i, \vec x_j) < r$).
    Let $y^{\text{(FT)}}$ be a FT classification task that is almost everywhere smooth on the manifold.

    Then, as PT dataset size (and thus the size of $G_\PT$) tends to $\infty$, and $r$ tends to zero, the local consistency of $y^{\text{(FT)}}$ over $G_\PT$ (and thus by Theorem~\ref{thm:nn_acc_to_LC} the nearest-neighbor accuracy of an SIPT embedder) will likewise tend to 1.
    
    \label{corollary:manifolds_deep}
\end{restatable}
Informally, these corollaries establish that when a shallow structural constraint is used (\eg a supervised classification objective), then the associated SIPT-equivalent model permits only minimal guarantees for FT performance, driven by the extent to which an FT task label is consistent within the classes under the supervised PT objective. In contrast, if a deep structural constraint is used, realized in Corollary~\ref{corollary:manifolds_deep} via $G_\PT$ being a nearest-neighbor graph over an arbitrary manifold $\mathcal M$, then a SIPT model permits a theoretical guarantee for FT performance that approaches unity as the pre-training dataset size grows for any FT task that is smooth over $\mathcal M$. 

In sum, this theoretical analysis shows that we can directly connect the structure induced in $\mathcal Z$ to downstream FT performance. As such, moving to new PT methods which leverage graphs $G_\PT$ with deeper structural constraints has the potential to markedly improve performance, as we will demonstrate on real-world datasets in our experiments. Complete proofs for all theoretical results and semi-synthetic experiments validating our theoretical findings in practice are in the Methods section.

\subsection*{Real-world Experiments: Datasets and Tasks}\label{sec:experiments}
We examine three data modalities for our experiments: \textsc{Proteins}, containing protein sequences; \textsc{Abstracts}, containing free-text biomedical abstracts; and \textsc{Networks}, containing sub-graphs of protein-protein interaction (PPI) networks.

In each data modality, we use different pre-training datasets and leverage different kinds of pre-training graphs $G_\PT$, test on publicly available benchmarks for FT tasks, and compare our SIPT methods to compelling baselines spanning both per-sample and/or per-token methods (Tables~\ref{tab:real_data_datasets}-\ref{tab:real_data_FT_tasks}). Further details on these aspects can also be found in the Methods Section.

\subsection*{Real-world Experiments: $\lossGGML$ and Training Procedures}
As discussed in the definition of our framework, a SIPT method differs from a standard PT method by (1) the choice of graph $G_\PT$ (Table~\ref{tab:real_data_datasets}) and (2) the design of the new, structure-inducing loss $\lossGGML$.
To define $\lossGGML$ in our experiments, we leverage ideas from \emph{structure-preserving metric learning} (SPML)~\cite{vert_supervised_2004,shaw_structure_2009,shaw_learning_2011}. SPML is a form of metric learning where positive relationships are defined by edges in a graph rather than a shared supervised label. 
We adapt two losses, a traditional contrastive loss~\cite{hadsell_dimensionality_2006} and a multi-similarity loss~\cite{wang_multi-similarity_2019}, from supervised metric learning to the graph-based, structure-preserving context of $\lossGGML$ terms in SIPT. 

In addition to these losses, in the \textsc{Abstracts} and \textsc{Proteins} domains, we use a warm-start procedure to initialize pre-training from existing language models rather than beginning from scratch. This saves significant computational time and allows for a powerful ablation study to isolate performance improvements to the introduction of our $\lossGGML$ term. Second, we perform extensive hyperparameter tuning studies on these two domains to identify appropriate values for $\lambdaGGML$, and adapt those findings to the \textsc{Networks} domain. Further details about the experimental setup, including formal statements of our contrastive and multi-similarity losses, are in the Methods section.

\subsection*{Result 1: Incorporating $\lossGGML$ performs comparably to or improves over all baselines across all 3 domains and 10 FT tasks}
To analyze our experiments, we compute the relative reduction of error
of the best performing SIPT model vs. the per-token or per-sample baselines across all FT tasks (Table~\ref{tab:main_results_table}). \emph{We can see that in 10/15 cases, SIPT improves over existing methods, and in no case does it do worse than either baseline.}
In some cases, the gains in performance are quite significant, with improvements of approximately 17\% (0.05 macro-F1 raw change) on AA, 6\% on SRE (0.01 macro-F1 raw change), and 4\% on RH (2\% accuracy raw change). \emph{SIPT models further establish a new SOTA on AA and RH and match SOTA on FL, ST, \& PF.}

We see in Figure~\ref{fig:graph_pt_res} how performance evolves over FT iterations for the \textsc{Networks} dataset to determine if the improvements observed at the final converged values are present throughout training. We see that SIPT methods converge faster to better performance than both baselines. Raw results across all settings are presented in the Methods section (Tables~\ref{tab:tape_FT_results}-\ref{tab:sciBERT_FT_results}).

\subsection*{Result 2: These performance gains are present across diverse modalities and pre-training graphs and outperform both per-sample and per-token baselines}
SIPT performance gains persist over all three data modalities and all different $G_\PT$ types we use here. This shows that explicitly regularizing the per-sample latent space geometry offers value across NLP, non-language sequences, and non-sequential domains, as well as while leveraging graphs including those defined by external knowledge, by self-supervised signals in the data directly, and by nearest-neighbor methods over multi-task label spaces. \emph{Furthermore, note that these improvements exist not only in comparison to standard language modelling approaches but also against existing methods that impose per-sample PT objectives, including single and multi-task classification objectives.}

\subsection*{Result 3: Observed gains are uniquely attributable to the novel loss $\lossGGML$}
As outlined in the Methods section, our experimental design permits us to determine how much of the observed gains in Table~\ref{tab:main_results_table} are due to the novel loss component, as opposed to, for example, continued training, new PT data, or the batch selection procedures used in our method which also indirectly leverage the knowledge inherent in $G_\PT$. Unsurprisingly, some gains are observed due to these other factors, and performance gains shrink when considering these ablation studies. However, even when comparing against the maximal performance baseline or ablation study overall, neither the direction of observed relationships nor the statistical significance of observed comparisons changes. \emph{Therefore, we can conclusively state that the performance improvements observed here are uniquely attributable to the new, structure-inducing components introduced by our framework.} Full ablation study results can be found in the Methods section (Tables~\ref{tab:tape_FT_results}-\ref{tab:sciBERT_FT_results}).

%% file: 030discussion.tex
We show that despite the breadth of research into PT methods, methods for imposing \emph{explicit} and \emph{deep} structural constraints over the per-sample, pre-training latent space $\mathcal Z$ are under-explored (Figure~\ref{fig:existing_pt_methods}). 
Our theoretical and empirical analyses \emph{show that this deficit matters in practice}. In particular, we define a new pre-training framework, \emph{structure-inducing pre-training} (SIPT), under which the PT loss is subdivided into two components: one which is designed to capture intra-sample (\eg per-token) relationships and one which is designed to constrain the per-sample latent space to capture relationships between samples given by a user-specified pre-training graph $G_\PT$. Under our framework, we show both theoretically and via experiments that the structure induced in $\mathcal Z$ can be directly connected to eventual fine-tuning performance. Empirically, we show that novel SIPT methods leveraging a variety of pre-training graphs can consistently outperform compelling existing PT methods across three real-world domains.

Our work highlights several important directions for future research. For example, are there losses better suited than metric learning losses for pre-training graphs---\eg, can we leverage the graph distance alongside the intra-batch distance to improve negative sampling strategies? In addition, can we produce theoretical results on convergence of pre-trained models? Can we advance the understanding of when and how pre-trained models converge to solutions that recover $G_\PT$? In a different direction, can pre-trained models reflect forms of structure beyond nearest neighbor relationships---\eg, such as by leveraging higher-order topological considerations or by matching a distance function rather than a discrete graph? We anticipate that further analyses of these and other questions will lead to new pre-training methods and enable pre-training to be successful across diverse domains.

%% file: 040figures.tex
\begin{figure}
    \centering
    \includegraphics[width=\linewidth]{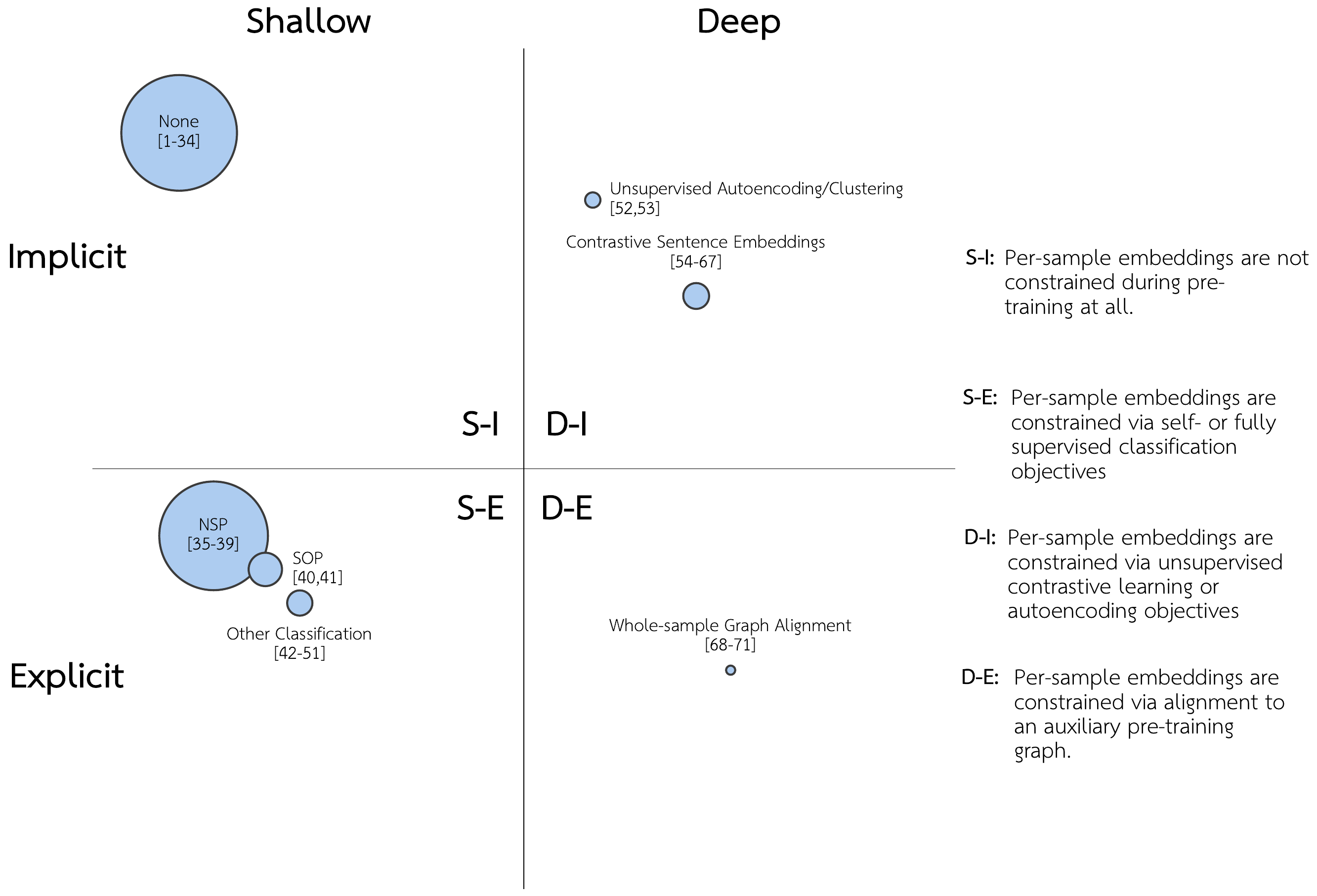}
    \caption{
    \textbf{Existing Pre-training (PT) Methods:}
    A summary of 71 existing natural language processing (NLP) and NLP-derived PT methods, categorized into clusters based on how they impose structural constraints over the PT (per-sample) latent space. Clusters are arranged on axes via manual judgements on whether the imposed constraint is \emph{shallow} vs. \emph{deep} and \emph{implici} vs. \emph{explicit}. Clusters are sized such that the area corresponds to the number of citations methods included in that cluster have received on average per month since first publication, according to Google Scholar's citation count.
    ``None'' captures models that leverage no pre-training loss over the per-sample embedding.
    ``NSP'' refers to ``Next-sentence Prediction,'' the per-sample PT task introduced in BERT~\cite{devlin_bert_2019}.
    ``SOP'' refers to ``Sentence-order Prediction,'' the per-sample PT task introduced in ALBERT~\cite{lan_albert_2019}. Note that over 90 studies in total were considered in our review, but only 71 met the inclusion criteria to be included in this figure. These methods are described in more detail in Methods Table~\ref{tab:existing_pt_methods} and in Appendix~\ref{sec:existing_LMPT_details}.
    }
    \label{fig:existing_pt_methods}
\end{figure}

\begin{figure}
    \centering
    \includegraphics{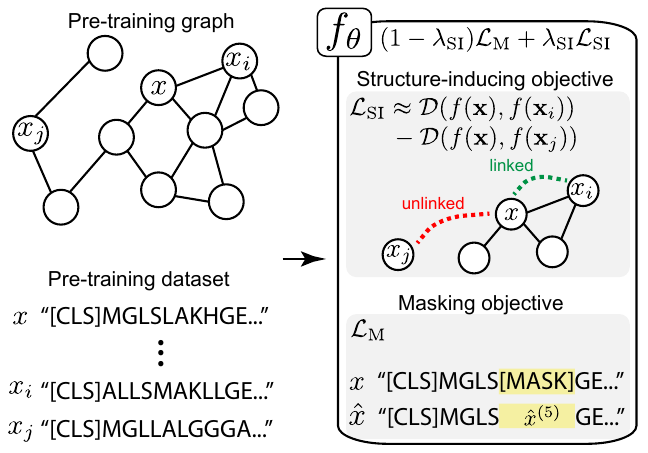}
    \caption{
    \textbf{Our Pre-training (PT) Framework}:
    We re-cast the PT formulation by taking a pre-training graph $G_\PT$ as an auxiliary input. $G_\PT$ is used to define a new structure-inducing objective $\lossGGML$, which pushes a pre-training encoder $f_{\vec \theta}$ to embed samples such that samples are close in the latent space if and only if they are linked in $G_\PT$.
    }
    \label{fig:framework_overview}
\end{figure}

\begin{figure}[t]
\includegraphics[width=0.8\linewidth]{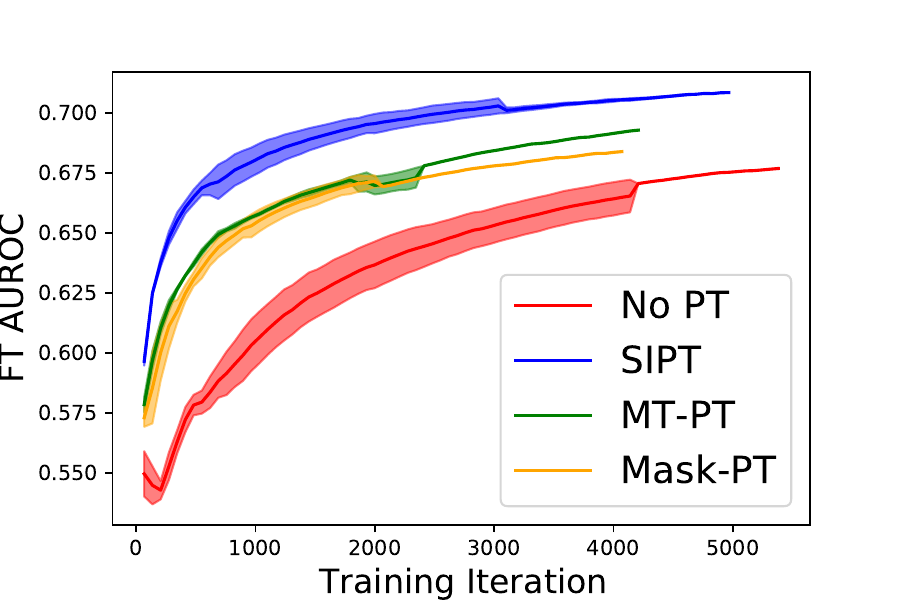}
\caption{
    \textbf{Fine-tuning (FT) Performance over \textsc{Networks}}:
    FT AUROC as a function of FT iteration for the \textsc{Networks} dataset. The SIPT method converges faster and performs better than intra-sample (masked node modelling) or per-sample (multi-task classification) pre-training.
}
\label{fig:graph_pt_res}
\end{figure}

%% file: 041methodsfigures.tex
\begin{figure}
    \centering
    \includegraphics[width=\linewidth]{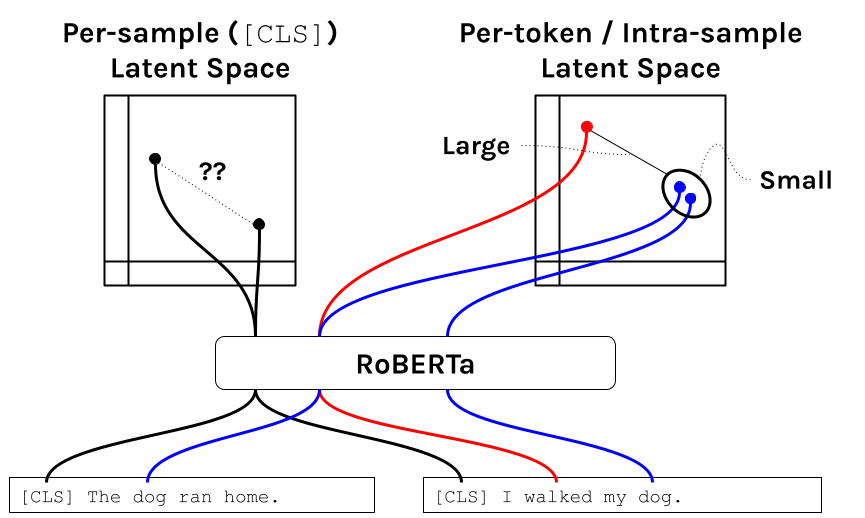}
    \caption{
    \textbf{Per-sample vs. Per-token Latent Space}
    Language model pre-training methods produce both per-sample and per-token latent spaces. Traditional language modelling objectives (illustrated here via the RoBERTa~\cite{liu_roberta_2019} model, which uses only a masked language model loss during pre-training) only constrain the per-token latent space.
    }
    \label{fig:BERT_intra_inter}
\end{figure}

\begin{figure*}
    \centering
    \includegraphics[width=\linewidth]{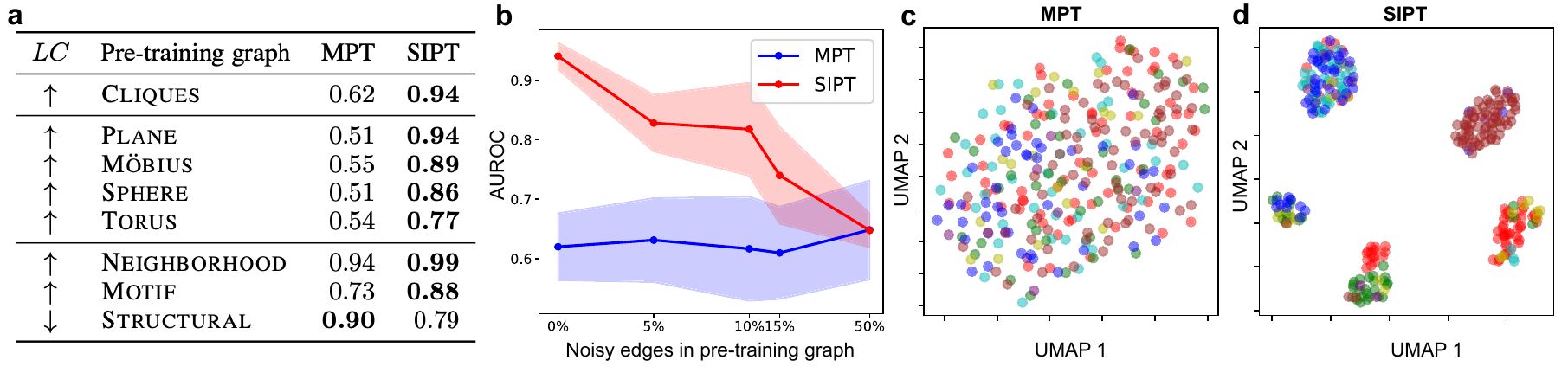}
    \caption{
    \textbf{Semi-synthetic Experiments Results:}
    \textbf{(a)} Comparisons between nearest-neighbor FT AUROC (higher is better) of \LMPTAbbr models and \methodAbbr models over various graphs with various forms of structural alignment. $LC$ indicates the label consistency between FT task and $G_\PT$ (Definition~\ref{def:LC}). \textbf{(b)} Nearest-neighbor FT AUROC vs.~noise rate. Up to 10\% noise \methodAbbr dramatically outperforms \LMPTAbbr, and at 50\% noise, the two approaches are equal. \textbf{(c-d)} Embedding space of MPT and SIPT models on the \textsc{M\"obius} dataset. Point colors indicate topic labels. SIPT's embedding space reflects the structure of the PT graph, whereas MPT does not.
    }
    \label{fig:table-noise-emb}
\end{figure*}

%% file: 042methodstables.tex
\begin{table}
    \centering
    \begin{tabular}{p{0.15\linewidth}p{0.22\linewidth}p{0.35\linewidth}p{0.25\linewidth}} \toprule
                            & \textsc{Proteins} & \textsc{Abstracts}       & \textsc{Networks}
        \\ \midrule
        Data Modality 
        ($\vec x_i$ is a...)& Protein Sequence & Biomedical Paper Abstract & PPI Network Ego-graph
        \\[25pt]
        PT Dataset          & Tree-of-life \mbox{\cite{zitnik_evolution_2019}}
                            & Microsoft Academic Graph \cite{wang_review_2019,hu_open_2020}
                            & \cite{GRAPH_PT_hu_strategies_2019} 
        \\[25pt]
        $G_\PT$: \mbox{$(\vec x_i, \vec x_j) \in E$ iff}
                            & $\vec x_i$ interacts with $\vec x_j$
                            & $\vec x_i$'s paper cites $\vec x_j$'s paper
                            & $\vec x_i$'s central protein agrees on all but 9 Gene Ontology (GO) labels with $\vec x_j$'s central protein.
        \\[25pt]
        Per-token \mbox{baseline}
                            & TAPE \mbox{\cite{rao_evaluating_2019}}
                            & SciBERT \mbox{\cite{beltagy_scibert_2019}}
                            & Attribute Masking \mbox{\cite{GRAPH_PT_hu_strategies_2019}}
        \\[25pt]
        Per-sample baseline & PLUS \mbox{\cite{PLUS_min_pre-training_2020}}
                            & None
                            & Multi-task learning \mbox{\cite{GRAPH_PT_hu_strategies_2019}}
        \\[25pt]
        FT Dataset          & TAPE \mbox{\cite{rao_evaluating_2019}}
                            & SciBERT \mbox{\cite{beltagy_scibert_2019}}
                            & \cite{GRAPH_PT_hu_strategies_2019}
        \\[25pt]
    \bottomrule\end{tabular}
    \caption{
    A summary of our datasets, tasks, and benchmarks. For example, for the \textsc{Proteins} domain, our pre-training dataset is the set of protein sequences contained in the tree-of-life dataset~\cite{zitnik_evolution_2019}, proteins are linked in our pre-training graph $G_\PT$ if and only if they interact according to the tree-of-life graph, and we compare over the fine-tuning tasks in the TAPE benchmark against both the raw, per-token baseline publicly available in the TAPE model~\cite{rao_evaluating_2019} as well as the per-sample baseline published in the PLUS pre-training model~\cite{PLUS_min_pre-training_2020}.
    }
    \label{tab:real_data_datasets}
\end{table}

\begin{table}
    \centering
    \begin{tabular}{llrrrrrr} \toprule
        \multirow{2}{*}{Domain}              & 
        \multirow{2}{*}{Task}                &
        \multicolumn{2}{c}{Vs. Per-Token PT} &
        \multicolumn{2}{c}{vs. Per-Sample}   \\ \cmidrule(lr){3-4} \cmidrule(lr){5-6}
               &      & RRE               & $\Delta$ & RRE               & $\Delta$ \\
        \midrule \multirow{5}{*}{\textsc{Proteins}}
               & RH   & \resB{7.0\%,1.2}  & \imp     & \resB{8.4\%,2.4}  & \imp     \\
               & FL   & \res{-0.8\%,1.3}  & \sam     & \resB{12.8\%,1.1} & \imp     \\
               & ST   & \resB{13.1\%,2.5} & \imp     & \res{2.2\%,2.8}   & \sam     \\
               & SS   & \resB{4.5\%,0.2}  & \imp     & \resB{4.5\%,0.2}  & \imp     \\
               & CP   & \resB{10.5\%}$^*$ & \imp     & \multicolumn{2}{c}{N/A}      \\
        \midrule \multirow{4}{*}{\textsc{Abstracts}}
               & PF   & \res{0.3\%,0.2}   & \sam     & \multicolumn{2}{c}{N/A}      \\
               & SC   & \res{2.4\%,4.1}   & \sam     & \multicolumn{2}{c}{N/A}      \\
               & AA   & \resB{17.7\%,6.5} & \imp     & \multicolumn{2}{c}{N/A}      \\
               & SRE  & \resB{6.7\%,0.4}  & \imp     & \multicolumn{2}{c}{N/A}      \\
        \midrule \textsc{Networks}
               &      & \res{7.8\%,5.2}   & \sam     & \res{5.1\%,2.7}   & \imp     \\
               
    \bottomrule \end{tabular}
    \caption{
    Relative reduction of error (RRE; defined to be $\frac{\text{[baseline error]} - \text{[$G_\PT$ model error]}}{\text{[baseline error]}}$) of models trained under our framework vs. published per-token or per-sample baselines. Higher numbers indicate models under our framework reduce error more and thus outperform baselines.
    The $\Delta$ column indicates whether the model offers a statistically significant improvement ($\uparrow$), no significant change ($\sim$), or a statistically significant decrease ($\downarrow$). Statistical significance is assessed via a $t$-test at significance level $p < 0.1$. Per-sample analysis and variance estimates for CP were infeasible due to the computational cost of this task.
    }
    \label{tab:main_results_table}
\end{table}

\begin{table}
    \centering 
    \begin{tabular}{p{0.15\linewidth}p{0.2\linewidth}lp{0.3\linewidth}l} \toprule
        FT Dataset & \multicolumn{2}{l}{FT Task} & Description & Metric \\ \cmidrule{2-3}
                   & Name & Abbr.                &             &        \\ \midrule
        \multirow{5}{0.2\linewidth}{TAPE \mbox{\cite{rao_evaluating_2019}}}
            & Remote Homology       &RH & Per-sequence classification task
                                            to predict protein fold category.   & Accuracy \\
            & Secondary Structure   &SS & Per-token classification task to
                                            predict amino acid structural
                                            properties.                         & Accuracy \\
            & Stability             &ST & Per-sequence, regression task to 
                                            predict stability.                  & Spearman's $\rho$ \\
            & Fluorescence          &FL & Per-sequence, regression task to 
                                            predict fluorescence.               & Spearman's $\rho$ \\
            & Contact Prediction    &CP & Intra-sequence classification 
                                            to predict which pairs of 
                                            amino acids are in contact
                                            in the protein's 3D conformation.    & Precision @ $L/5$ \\ \midrule
        \multirow{4}{0.2\linewidth}{SciBERT \mbox{\cite{beltagy_scibert_2019}}}
            & Paper Field           &PF & Per-sentence classification problem 
                                          to predict a paper's area of study
                                          from its title.                       & Macro-F1 \\
            & SciCite               &SC & Per-sentence classification problem to predict citation intent & Macro-F1 \\
            & ACL-ARC               &AA & Per-sentence classification problem to predict citation intent & Macro-F1 \\
            & SciERC                &SRE& Per-sentence relation extraction & Macro-F1 \\ \midrule
        \textsc{Networks} \mbox{\cite{GRAPH_PT_hu_strategies_2019}}
            &                       &   & Multi-label binary classification into 40 Gene Ontology terms & Macro-AUROC \\
    \bottomrule \end{tabular}
    \caption{
    Fine-tuning tasks.
    }
    \label{tab:real_data_FT_tasks}
\end{table}

%% file: 050online_methods.tex
\newcommand{\abs}[1]{\left|#1\right|}

\subsection*{Per-token vs. Per-sample Latent Space: Definition of $\mathcal Z$}
\label{methods_subsec:definition_of_Z}
Let $f_{\vec \theta}$ be a pre-training (PT) model trained over a dataset $\mat X \in \mathcal X^{N_\PT}$. Furthermore, let us assume that samples $\vec x \in \mathcal X$ are composed of smaller units (\eg tokens, sequence time-points, nodes in a network, etc.). Let us denote this by saying that $\vec x = w_1, w_2, \ldots, w_{n_{\vec x}}$. Finally, as is true in natural language processing (NLP) and NLP-derived settings, we assume that $f_{\vec \theta}$ can be seen to produce output embeddings for both the entire sample $\vec x$---which we will denote by $f_{\vec \theta}(\vec x)$---and for the internal tokens individually---which will denote by $f_{\vec \theta}(w_j | \vec x)$. For example, in the BERT model~\cite{devlin_bert_2019}, $f_{\vec \theta}(\vec x)$ will be given by the output embedding of the \texttt{[CLS]} token of $\vec x$ and $f_{\vec \theta}(w_j | \vec x)$ will be given by the output embedding of the $j$-th token in $\vec x$.

We can then formally define the per-sample latent space, $\mathcal Z^{(\text{S})}$ (which we will also refer to as $\mathcal Z$ without the superscript), and the per-token (aka intra-sample) latent space $\mathcal Z^{(\text{T})}$ (Definitions~\ref{def:per_sample_latent_space}~\&~\ref{def:per_token_latent_space}, and Figure~\ref{fig:BERT_intra_inter}).
\begin{definition}{Per-Sample Latent Space}
We define the \emph{per-sample latent space} induced by $f_{\vec \theta}$ as $\mathcal Z^{(\text{S})} = \{f_{\vec \theta}(\vec x) | \vec x \in \mathcal X\}$. We will also use $\mathcal Z$ with no superscript to refer to this space.
\label{def:per_sample_latent_space}
\end{definition}
\begin{definition}{Per-token/Intra-sample Latent Space}
We define the \emph{per-token latent space} (also known as the \emph{intra-sample latent space}) induced by $f_{\vec \theta}$ as $\mathcal Z^{(\text{T})} = \{f_{\vec \theta}(w_j | \vec x) | \vec w_j \in \vec x, \vec x \in \mathcal X\}$.
\label{def:per_token_latent_space}
\end{definition}

Both of these spaces are very different and are useful in different contexts; for a task like named entity recognition, where the unit of classification is a single or short span of tokens, analyzing the per-token latent space will be more informative, whereas for a task like sentiment analysis, where the unit of classification is an entire sample (sentence), the per-sample latent space would be preferred~\cite{devlin_bert_2019}. Furthermore, another key difference between these spaces is that the traditional PT language model objective only induces significant constraints on the geometry of the per-token latent space and does not impact the per-sample latent space at all. This illustrates a gap in the capabilities of PT methods. In our work, we are concerned with precisely this gap and focus our attention on $\mathcal Z$ (\ie $\mathcal Z^{\text{(S)}}$). We focus our attention on the per-sample latent space for 3 reasons:
\begin{enumerate}
    \item There has been significantly more research on how to regularize the per-token latent space than the per-sample latent space, as we show in our extensive review (Table~\ref{tab:existing_pt_methods}).
    \item In many domains outside of NLP, the per-sample latent space is often of much greater interest than the intra-sample latent space. For example, in modelling protein sequences~\cite{rao_evaluating_2019}, drug structures~\cite{GRAPH_PT_hu_strategies_2019}, or electronic health record time series~\cite{mcdermott_comprehensive_2020}, per-sample tasks are of much greater interest than intra-sample tasks.
    \item Even within NLP, modern methods struggle much more with representing whole passages of text rather than short, isolated spans. This is evidenced by the battery of work examining sentence representations atop pre-trained language models~\cite{BERT_FLOW_li-etal-2020-sentence,SIMCSE_gao-etal-2021-simcse}.
\end{enumerate}

\subsection*{Why is NLP Different than Other Domains?}
\label{methods_subsec:nlp_vs_other}

In this work, we have implicitly argued that because a PT objective like masked language modelling (MLM) will not necessarily directly enrich the per-sample latent space $\mathcal Z^{(\text{S})}$, it may yield models less well suited to downstream per-sample tasks than other approaches.
One seeming contradiction to this is that methods in NLP like RoBERTa~\cite{liu_roberta_2019} (for which MLM is the only PT objective) succeed across both per-token and per-sample tasks. 

In fact, this observation does not contradict our hypothesis but reflects a unique advantage of the natural language modality that does not apply in other domains. In particular, in the NLP domain (and not in other domains), we can leverage the flexibility of the language to sidestep any deficit in $\mathcal Z^{(\text{S})}$ by re-framing per-sample tasks as per-token, language modelling tasks. 
Significant literature exists documenting this phenomenon through the lenses of prompting, cloze-filling models, text-to-text transformers, and  theoretical analyses~\cite{GPTIII_brown2020language,saunshi2021a,T0pp_sanh2021multitask,T5_JMLR:v21:20-074,UNI_LM_NEURIPS2019_c20bb2d9}.
For example, \cite{saunshi2021a} examines the efficacy of pre-trained language models on sentiment analysis explicitly and show that the language modelling component alone can be used in a per-token manner to indirectly solve a review sentiment analysis task by judging the likelihood of following the review with a ``:)'' emoji vs. a ``:('' emoji. In this way, they shift the \emph{per-sample} task of sentiment analysis to a \emph{per-token} task via the (inserted) emoji.

However, language model pre-training has also inspired many derived methods to be used in other non-NLP domains. For example, in modelling graphs, \cite{GRAPH_PT_hu_strategies_2019} has examined vertex or edge-masking strategies reminiscent of MLM, with vertices and edges analogous to tokens and entire graphs whole samples; in modelling time series data, \cite{mcdermott_comprehensive_2020} has examined masked imputation models, with timepoints analogous to tokens and whole time series to samples; and in modelling protein sequences, \cite{PLUS_min_pre-training_2020} has used masked language modelling directly, with individual amino acids representing tokens and entire proteins representing samples. \emph{In all three of these domains, we cannot re-frame per-sample tasks as ``per-token'' tasks as we can in NLP, and accordingly, the problem of insufficient per-sample latent space regularization will likely be much more severe in these domains. Accordingly, existing work, including the three works referenced above, all find that augmenting the language model pre-training task with additional, per-sample level supervised tasks can be beneficial, or even absolutely essential, to improving performance~\cite{GRAPH_PT_hu_strategies_2019,yoon_vime_2020,mcdermott_comprehensive_2020,PLUS_min_pre-training_2020}.}

\subsection*{Pre-training Review Methodology}
\label{methods_subsec:details_on_review}
Papers were selected via a manual search of the natural language processing (NLP) and NLP-derived pre-training methods (\ie, methods focused primarily on other domains or on multi-modal domains were excluded) via Google Scholar as well as by crawling through references of papers already included. Citation counts for each work were obtained via Google Scholar on August 2nd, 2022. Publication date (used to calculate citations per month since publication date) was computed as the earlier of either (1) the paper's venue-specific date of publicatoin or (2) the first submission date to the arXiv or BioRxiv platform, as referenced via an exact title match. A manual review was done to classify how pre-training methods constrain latent space geometry and assign subjective, numerical ``shallow-deep'' and ``explicit-implicit'' axes scores. In total, over 90 methods were examined, of which 71 were suitable for inclusion in numerical review results (Figure~\ref{fig:existing_pt_methods} and Table~\ref{tab:existing_pt_methods}). All methods considered are summarized and categorized (and reasons for exclusions are given) in Appendix~\ref{sec:existing_LMPT_details}.

\subsection*{Further Analysis of Reviewed Methods}
This work has extensively examined how existing pre-training methods constrain the \emph{per-sample} latent space. However, it is also worth examining how these methods constrain the per-token latent space to demonstrate the extent to which per-sample objectives are under-explored in current pre-training research. To that end, we break down all of the studies included in our review not only by how they constrain their per-sample latent spaces but also by how they constrain their per-token latent spaces (Table~\ref{tab:existing_pt_methods}). These groupings are also done at a greater granularity than the previously examined categories to offer more insight into which methods use which techniques. We see that not only are there more types of per-token latent space constraints leveraged (10 vs. 7), but also methods consistently leverage a much greater diversity of per-token constraints vs. per-sample constraints (1.45 per-token constraints per method vs. 0.58 per-sample constraints, on average). We can further see from Figure~\ref{fig:existing_pt_methods} that the citation volume for works in this space is also heavily concentrated around methods that first employ no per-sample PT objective, followed by methods that only impose shallow, explicit methods, which further establishes this research gap.

\include{051pt_methods_table}

\subsection*{Constraints on $\lossGGML$ in our Framework}
\label{methods_subsec:lossGGML_constraints}
Formally, for $\lossGGML$ to be valid, then there must exist a distance function $d: \mathcal Z \times \mathcal Z \to \R$, radius $r \in \R$, and loss value $\ell^* \in \R$ such that at any solution $\vec \theta^*$ for which $\lossGGML(\vec \theta^*) < \ell^*$, the learned embeddings $\vec z_i = f_{\vec \theta^*}(\vec x_i)$ must recover the graph $G_\PT$ under a radius nearest neighbor connectivity algorithm via distance function $d$ and radius $r$---\ie, it must be the case that $(\vec x_i, \vec x_j) \in E$ if and only if $d(f_{\vec \theta^*}(\vec x_i), f_{\vec \theta^*}(\vec x_j)) < r$. Furthermore, for the particular graph $G_\PT$ and latent space $\mathcal Z$, the set of $\vec \theta^*$ such that $\lossGGML(\vec \theta^*) < \ell^*$ must be non-empty (\ie such a solution must exist).

\subsection*{Realizing Existing Methods in our Framework}
\label{methods_subsec:existing_methods}

Let $\mat X \in \mathcal X^{N_\PT}$ be the pre-training dataset throughout this section. In cases where we have some auxiliary information (\eg, supervised, per-sample, pre-training labels), they will be denoted by $\mat Y \in \mathcal Y^{N_\PT}$.

\xhdr{Methods with no per-sample objectives}

Naturally, we can realize any method that only employs a per-token pre-training objective within our framework simply by setting $\lambdaGGML = 0$. This realization is trivial and offers no insight into the suitability of these pre-training methods for downstream per-sample tasks.

\xhdr{Methods with a supervised, single-task per-sample objective (\eg, BERT~\cite{devlin_bert_2019})}

A simple, single-task, per-sample, classification pre-training objective induces a geometric constraint in the output latent space on the basis of the inner product ``distance'' between samples of the same vs. different class labels. 
We can use this observation to realize a reduction from a valid SIPT objective to the original classification objective. In particular, we can introduce a graph $G = (\{\vec x_i \in \mat X\}, \{(\vec x_i, \vec x_j) | y_i = y_j\})$ which consists of cliques corresponding to each unique label $c \in \mathcal Y$. Then, leveraging any structure-preserving metric learning loss with a cosine distance objective will, at optimality, recover a solution that also satisfies the original classification objective, where we use centroids of the induced clique embeddings to represent class embeddings.

\xhdr{Methods with a supervised, multi-task per-sample objective (\eg, MT-DNN~\cite{MT_DNN_liu_multi-task_2019})}

A slightly more complicated case is when methods employ a multi-task, per-sample classification objective. In this case, there are two ways to realize this task within the SIPT framework. First, we can simply transform the multi-task objective into a single-task objective by constructing a new label-space consisting of the Cartesian product of all label spaces for each task individually. This will greatly increase the number of ``labels'' in the task, but then the problem can be realized via a graph of disconnected cliques much like in the single-task setting. 

However, there is another manner in which we can realize this objective in the SIPT framework; In particular, suppose our collection of tasks consists of $k$ label spaces: $\mathcal Y = \mathcal Y_1 \times \cdots \times \mathcal Y_k$. Then, we can construct a graph $G = (V, E)$ such that:
\begin{enumerate}
    \item the vertices consist of all pre-training samples $\vec x_i$ as well as auxiliary nodes corresponding to each label $c_h^{(j)} \in \mathcal Y_h$ across each task: $V = \{\vec x_i \in \mat X\} \cup \mathcal Y_1 \cup \cdots \cup \mathcal Y_k$
    \item the edges contain links between each sample $\vec x_i$ and label $y_h^{(i)}$ across all tasks $1 \le h \le k$: $E = \{(\vec x_i, c_h^{(j)}) | y_h^{(i)} = c_h^{(j)}\}$. 
\end{enumerate}

Then, we can see that if we solve the SIPT problem under a structure-preserving metric learning loss, we will naturally have produced embeddings for each $\vec x_i$ which are close (in inner-product distance space) to the class embeddings corresponding to their labels for each task, while they are also far from other, non-matching class embeddings, as desired.
This second approach is more useful to us in considering the ramifications of this style of constraint because it enables us to make more rigid theoretical guarantees via the SIPT theory.

\xhdr{Methods with a based contrastive per-sample objective (\eg, GraphCL~\cite{GraphCL_NEURIPS2020_3fe23034}})

It is challenging to realize contrastive learning approaches within the SIPT framework, but it is still possible. Here, we highlight two distinct types of contrastive learning approaches we can capture within SIPT: a noising/augmentation-based approach, in which sample embeddings are constrained to be similar to embeddings of noised versions of said samples; and a multi-modal (or multi-lingual) contrastive approach, in which there exists a 1:1 mapping between two different sub-modalities within $\mat X$ which is used to join those two modalities into a unified latent space (\eg a model which constrains embeddings of English sentences to be close to embeddings of their french translations, but far from unrelated sentences).

To consider the augmentation/noising policy type first, let $h: \vec x_i \mapsto \Tilde{\vec x_i}$ represent the noising transformation. Then, to build an analogous SIPT model to this model, we construct an augmented dataset consisting of all original data points alongside all possible transformed versions of the original data points under $h$: $\mat X' = \mat X \cup \left(\bigcup_{i=1}^{N_\PT} \mathrm{Im}\left(h\middle\vert_{\vec x_i}\right)\right)$. Note that even in contexts where $h$ is continuous (and thus has an infinite image), we can still construct this dataset in practice because training is only performed over a finite number of steps, meaning our augmented dataset $\mat X'$ need only be expanded to cover a finite number of augmentations. Then, the associated pre-training graph is simple; we simply use every sample in the augmented dataset $\mat X'$ as a vertex and connect any two samples if and only if one is a transformed version of the other. This forms a graph of many disconnected stars (one star for each original datapoint $\vec x_i$), and thus it does not directly enforce any particular geometry via our current theory. However, in cases where dataset size is sufficiently large, $h$ sufficiently expressive, and data density sufficiently high, then the natural continuity of any neural network model will induce additional, auxiliary connections across these stars (if, for example, the noised versions of two distinct samples have a high probability of being very similar), which increases the depth of the geometric constraints enforced. Quantifying the exact parameters of these interactions, however, we leave to future work.

In the case of the multi-modal/multi-lingual contrastive alignment objective across $k$ modalities, our setup is much simpler: we simply let $G_\PT$ be a $k$-partite graph whose samples consist of individual data points (across all modalities) and edges connect samples that compose a matching pair across modalities (\eg edges link English sentences to their french translations). The extent to which this constrains the output geometry in practice, then, comes down to several questions: (1) Is the cross-modal alignment a one-to-one, one-to-many, or many-to-many alignment (which impacts the geometry of the resulting graph), (2) How large and dense is the dataset (which impacts the extent to which additional, indirect edges will be induced due to continuity in practice), and (3) How do other pre-training objectives constrain the individual modalities separately? In a case where this graph is one-to-one, and no other constraints are induced in each modality separately, this objective will offer only minimal constaints as the resulting graph will consistent of many disconnected 2-cliques.

\xhdr{Methods with a per-sample graph-alignment objective (\eg, KEPLER~\cite{wang_kepler_2019})}

Methods that explicitly align samples with a provided pre-training graph (KEPLER~\cite{wang_kepler_2019}, CK-GNN~\cite{CK_GNN_fang2021knowledge}, XLM-K~\cite{XLM_K_jiang2022xlm}, and WebFormer~\cite{WEBFORMER_guo2022webformer}) are naturally already realized within SIPT, so need no further commentary here.

\subsection*{Structure-inducing Losses Examined in this Study}
\label{methods_subsec:framework_losses}

\xhdr{Multi-similarity loss}

The multi-similarity loss, parametrized by $w_+$, $w_-$, and $t$, is given below:
\begin{align*}
    \lossGGML &=
        \frac{1}{N w_+}\log\(
            1 + \!\!\!\! \sum_{(i, j) \in E} e^{-w_+ \(\ip{f_{\vec\theta}(\vec x_i), f_{\vec\theta}(\vec x_j)} - t\)}
        \) + \frac{1}{N w_-}\log\(
            1 + \!\!\!\! \sum_{(i, j) \not \in E} e^{w_- \(\ip{f_{\vec\theta}(\vec x_i), f_{\vec\theta}(\vec x_j)} - t\)}
        \),
\end{align*}

\xhdr{Contrastive loss}

Our contrastive loss is modeled after \cite{hadsell_dimensionality_2006}'s version. For this loss, we assume we are given the following mappings: `$\mathrm{pos}$', which maps $\vec x$ into a positive node (\ie, linked to $\vec x$ in $G_\PT$), and `$\mathrm{neg}$', which maps $\vec x$ into a negative node (\ie, not linked to $\vec x$ in $G_\PT$). The union of a seed minibatch $B$ of points $\mat X_B$ and its images under `$\mathrm{pos}$' and `$\mathrm{neg}$' mappings form a full minibatch. 
This loss is specified by the positive and negative margin parameters $\mu_+$ and $\mu_-$ as:
\begin{align*}
    \lossGGML^{(\text{CL})} &=
     \frac{1}{N}\sum_{\vec x_i \in \mat X} \max(\mathcal D(\vec x_i, \mathrm{pos}(\vec x_i)) - \mu_+, 0) + \frac{1}{N}\sum_{\vec x_i \in \mat X} \max(\mu_- - \mathcal D(\vec x_i, \mathrm{neg}(\vec x_i)), 0).
\end{align*}

\subsection*{Additional Choices within the SIPT Framework}
In addition to a loss term, we can use negative sampling to improve efficiency. Using the full graph $G_\PT$, which is not available in many contexts where negative sampling is employed, we can leverage the distance between samples calculated on $G_\PT$, which provides a complementary source of information beyond embedding space distance alone. For example, one could use this to limit negative samples within the same connected component, but more complex strategies based on graph sampling (\eg~\cite{Zeng2020GraphSAINT:}) could also be used.

\subsection*{Proof of Theorem~\ref{thm:nn_acc_to_LC}}
We begin by defining the notion of ``Local Consistency,'' which (informally) quantifies how ``smooth'' a given fine-tuning task label is over a graph $G_\PT$ (Definition~\ref{def:LC}). In addition, note that throughout all proofs, we will assume that the PT and FT datasets are iid, that FT tasks, though they may be unobserved over PT samples, are well defined over the entire PT and FT domain and thus true labels do exist (though they may be unknown) for PT samples, and that the sampling distribution of the PT/FT data has full support over the label-space of any considered task.
\begin{definition}[Local Consistency]
Let $y: X \to \mathcal Y$ be a task over a domain $X$, and let $G = (V, E)$ be a graph such that $X \subseteq V$. The \emph{local consistency} $\mathrm{LC}_G(y)$ is the probability that a node's label $y(x)$ agrees with the majority of labels of $x$'s neighbors in $G$:
\[\mathrm{LC}_G(y) = \mathbb{P}\(y(x) = \argmax_{c \in \mathcal Y} \sum_{x' \in X | (x, x') \in E} \mathds{1}_{y(x') = c}\).\]
Note this is closely related to \emph{homophily}~\cite{zhu2020generalizing,huang2020graph,zhang2016homophily}.

\label{def:LC}
\end{definition}
With Local Consistency defined, we can now formally prove Theorem~\ref{thm:nn_acc_to_LC}, reproduced below.
\thmOne*
\begin{proof}
Given $f$ realizes \methodAbbr-optimal embeddings, we know that if we define a $r$-NN predictor via the same radius $r^*$ at which $f$ achieves optimality, then this predictor will be correct exactly as often as the label of a given node in the graph $G_\PT$ agrees with the labels of its neighbors---which is  $\mathrm{LC}_{G_\PT}(y)$. This classifier may not be well defined for small FT dataset sizes. However, as if data is not sufficiently dense, there may be no data points within the radius $r$ of a given query. Similarly, without sufficient PT data, the $\mathrm{LC}$ computed over the empirical distribution of the graph $G_\PT$ may be a poor proxy for the true distribution. As PT and FT dataset sizes increase, however, we can achieve at least this performance. We may be able to achieve even higher performance if other effects motivate stronger performance at radii smaller than $r^*$, but this is not guaranteed.
\end{proof}

\subsection*{Proof of Corollary~\ref{corollary:cliques_shallow}}
\coroCliques*
\begin{proof}
    This follows directly from the definition of Local Consistency, $G_\PT$, and the law of total probability.
    In particular,
    \begin{align*}
        \mathrm{LC}_{G_\PT}(y_\FT)
          &= \mathbb{P}\(
            y_\FT(\vec x_i) = \argmax_{\ell \in \mathcal Y_\FT} \sum_{\vec x_j \in \mat X_\PT | (\vec x_i, \vec x_j) \in E(G_\PT)} \mathds{1}_{y_\FT(\vec x_i) = \ell}
          \) \\
          &= \mathbb{P}\(
            y_\FT(\vec x_i) = \mathrm{MC}(\vec x_i, y_\FT)
          \) \\
          &=\sum_{\ell_\PT \in \mathcal Y_\PT} \mathbb{P}(y_i = \ell_\PT) \mathbb{P}(y_\FT(\vec x_i) = \mathrm{MC}(\vec x_i, y_\FT) | y_i = \ell),
    \end{align*}
    With Local consistency found, a simple application of Theorem~\ref{thm:nn_acc_to_LC} completes the proof.
\end{proof}
Note that this has a dependence on the PT dataset size as the probabilities $\mathbb{P}$ are taken over the empirical distribution induced by the dataset $\mat X_\PT$ and graph $G_\PT$ inherent in local consistency --- if $\mat X_\PT$ is too small, these empirical distributions will be poor proxies for the true distribution and this bound will not hold tightly. However, once saturation is reached, it will not improve beyond this fixed upper bound relating to task correlation.

\subsection*{Proof of Corollary~\ref{corollary:manifolds_deep}}
\coroManifolds*
\begin{proof}
    As $r \to 0$, provided PT dataset size increases at a sufficient associated rate so as to maintain a constant minimum degree of $G$, we have the property that the total diameter over $\mathcal M$ contained in a node's local neighborhood within $G_\PT$ likewise decreases. Given some fixed node $\vec x \in \mathcal M$ that is within the interior of a set of constant $y_\FT$ label, this implies that, eventually, it will grow sufficiently small that all of $\vec x$'s neighbors share the same label as $\vec x$ under $y_\FT$. 
    
    More concretely, it is clear that this point will occur exactly when $r$ is the geodesic distance between $\vec x$ and the boundary of the surrounding constant-label patch containing $\vec x$. But, it is clear that the only sections of $\mathcal M$ will not have the property that neighborhoods around points will be constant w.r.t. $y_\FT$ labels will almost everywhere be patches within distance $r$ of the points where $y_\FT$ changes.
    
    This implies that as $r \to 0$, then almost everywhere will the neighborhoods around a node $\vec x$ be constant w.r.t. $y_\FT$. However, this implies that almost everywhere would $y_\FT$ display perfect local consistency, as desired.
\end{proof}

\subsection*{Semi-synthetic Experiments Validating Theoretical Results}\label{methods_sec:synthetic_experiments}

We can further validate the theoretical analyses of our framework via semi-synthetic experiments. In particular, we create several datasets of natural language sentences augmented with synthetic graphs with known relationships to certain FT tasks (e.g., low or high local consistency, low or high rates of noise). We then use these datasets to validate three important properties of PT methods:
First, do PT methods trained with a $\lossGGML$ and $G_\PT$ yield \notzeroshot FT performance in accordance with our theory? In particular, do (a) FT tasks with high local consistency over the PT graph offer better performance, and (b) those with very low local consistency offer worse performance?
Second, do PT methods trained with a $\lossGGML$ and $G_\PT$ suffer significantly when pre-training graphs are polluted with noise?
Finally, third, do the latent space geometry regularizing properties of $\lossGGML$ yield methods whose embeddings more clearly cluster than embeddings produced by traditional pre-training alone?

\subsubsection*{Pre-training \& fine-tuning datasets}
Across all experiments, our synthetic datasets consist of free-text sentences from \url{https://www.kaggle.com/mikeortman/wikipedia-sentences} (CC BY-SA 4.0 License).

Topics were assigned to these sentences by running Latent Dirichlet Allocation via Scikit-learn~\cite{scikit-learn} over a Bag-of-words representation to 100 topics, with otherwise default parameters. Given the probabilities over all 100 topics, we treated the prediction of the most probable topic as a 100-class multi-class classification problem for our FT task in these experiments.

To test across various graphs, we produce a number of pre-training graphs per experiment, as detailed below. 

\subsubsection*{Pre-training graphs}
We use graphs spanning 3 categories. (1) A graph (\textsc{cliques}) consisting of disconnected cliques, where sentences are linked in the graph if they share the same topic label. (2) Graphs composed of nearest-neighbor graphs defined over simplicial manifolds built using topic probabilities to localize sentences onto simplices. We explore manifolds with a range of topological complexity, including: \textsc{Plane}, \textsc{M\"obius}, \textsc{Sphere}, and \textsc{Torus}. Finally, (3) we define three graphs according to a mechanistic process that allows us to control how topic labels relate to graph structure: first, so that topics are maximally conserved within local neighborhoods (\textsc{Neighborhood}); second, by assigning sentences to nodes in the graph such that each graph motif corresponds to a unique topic (\textsc{Motif}); and third, such that node topics are driven by non-local graph structural features, on the basis of graphlet degree vectors  (\textsc{Structural}). Details for each pre-training graph formation are given below.

\xhdr{\textsc{Cliques} Graph Setup}

To construct the Cliques graph setting, we choose a random subset of sentences as $\mat X_\PT$ and define $G_\PT = (\mat X_\PT, E)$ such that $(\vec x_i, \vec x_j) \in E$ if and only if $\vec x_i$ and $\vec x_j$ share the same topic label.

\xhdr{\textsc{Plane}, \textsc{M\"obius}, \textsc{Sphere}, \& \textsc{Torus} Graphs}

For these graphs, we take a more involved practice to localize sentences onto specifiable simplicial manifolds, then construct pre-training graphs via radius nearest neighbor graphs on those manifolds. This involves several steps:
\begin{description}
    \item[Localizing Sentences on Simplices]
    We can localize any sentence in our overall dataset onto a 2-simplex by mapping them onto the (re-normalized) probabilities associated with their top-3 topics. Doing this means that the simplex on which they are localized has vertices corresponding to possible topics among our 100 total topics.
    \item[Stitching Topic-simplices Into Manifolds]
    Given these topic-simplex localized sentences, we need to construct our manifolds. To do so, we first produce any arbitrary simplicial tiling of a 2-manifold. With this tiling, all that remains to localize sentences onto the manifold is to find a self-consistent mapping of topics to simplex vertices (in the tiling) such that all topic-simplices induced by this mapping have sufficiently many associated samples to enable roughly uniform sampling.
    \item[Sampling Points]
    After finding a self-consistent map of topics to simplicial tiling vertices that satisfy density requirements, we can sample sentences onto the manifold. To make this process more uniform, we also calculate the relative entropy of each sentence (over the re-normalized probabilities of the top-3 topics), bin those entropies into buckets, then sample first what entropy bucket we wish to draw from such that the induced distribution of sentence entropies is approximately uniform, then sample within that entropy bucket. 
    \item[Calculating on-Manifold Distances]
    Finally, with sentences sampled and localized onto a simplicial manifold, we then need to compute approximate geodesic distances to enable building radius-nearest-neighbor graphs over these sentences. To do so, we use an approximate algorithm that considers only on-simplex distance (\eg, it does not consider any curvature penalties) which is equivalent to calculating the distance between any pair of points over the simplices presuming they were flattened onto a plane (this flattening naturally does not preserve manifold topology, but along only the shortest path between any particular set of two points it is always possible to do so with a 2-manifold).
\end{description}

The above process describes how to produce a radius-nearest-neighbor graph for any specifiable manifold using our topic-model outputs. We do this for simplicial manifolds that correspond topologically to a simple plane (\textsc{Plane}), a m\"obius strip (\textsc{M\"obius}), a sphere (\textsc{Sphere}), and a torus (\textsc{Torus}).

\xhdr{\textsc{Structural}, \textsc{Neighborhood} \& \textsc{Motifs} Graphs}

In order to form these examples, we must (1) define our overall graphs, (2) featurize these graphs in a manner that is reflective of different forms of graph structure, then (3) use these featurizations to assign sentences to graph nodes to form our pre-training dataset. 
\begin{description}
    \item[Graph Construction] We sample graphs by first building a base cycle of a parametrized size, then add motifs along this cycle by sampling small graphs from all possible connected graphs of size less than 6 nodes.
    \item[Node Featurization] Nodes in this graph are then assigned internal features based on three notions of graph topology. For the ``Neighborhood'' label, a node $n$ is identified according to an index-vector indicating which nodes in the graph are within shortest-path distance 3 of $n$. For the ``Motif'' label, $n$ is identified based on its membership either in the base cycle or any of the attached random subgraphs. For the ``Structural'' label, $n$ is identified based on its graphlet degree vector (of order 4). For structural and homophily features, categorical labels are then produced by feeding these raw representations through a $k$-means clustering algorithm.
    \item[Sentence Assignment] We assign sentences to nodes in multiple ways so that we can produce datasets that reflect each of the notions of graph structure discussed previously. In particular, for either the neighborhood, motif, or structural labels, each sentence topic is matched to a node label, then sentences are assigned randomly to nodes in the graph with a matching topic label. Note that this produces a dataset where the graph structure is only partially reflected by the node's features, which is itself another useful test of the \methodAbbr method, as it would not be useful if \methodAbbr could only capture data in contexts where the graph was perfectly reflected by the node features themselves.
\end{description}

\xhdr{Expected local consistency between graphs $G_\PT$ and the topic prediction FT task} 

Of all these graphs, we expect that topics will display a low local consistency over the \textsc{Structural} graph and a moderately high local consistency over the \textsc{Motif} graph (as graph motifs are all connected components), and high local consistency everywhere else. 

\subsubsection*{Network Architecture \& Hyperparameters}

The Cliques and Mechanistic experiments use a shallow Transformer model with 2 layers and 10 hidden units. The Manifold experiments use a 3-layer Transformer model with 256 hidden units. Hyperparameters were not tuned but were chosen by hand to produce as small a network as possible while permitting reasonable learning dynamics.

\subsubsection*{Experimental setup}
To answer our three questions, we will pre-train models under both traditional LM pre-training alone and a new, structure-inducing PT (SIPT) method within our paradigm that augments the loss with a contrastive learning loss over $G_\PT$, with $\lambdaGGML=0.1$. Both models use a shallow transformer encoder for $f_{\vec \theta}$ and a character-level tokenization scheme. Final results are reported via the AUROC of 3-nearest-neighbor classifiers over the latent space, per-sample embeddings. In line with our theoretical predictions, we expect to see higher NN FT performance in all settings where the FT task (topic prediction) has high local consistency over the graph $G_\PT$ (all graphs except \textsc{Structural}) and worse performance in the case where the local consistency is very low (\textsc{Structural}). 

We also assess the stability of our method as the graph $G_\PT$ is noised using the \textsc{Cliques} graph by randomly adding additional edges with varying rates.

\subsubsection*{Semi-synthetic Result 1: \methodAbbr improves performance over \LMPTAbbr by $0.26 \pm 0.13$ AUROC on graphs where the topic task has a high local consistency}
As can be seen in Figure~\ref{fig:table-noise-emb}a, \methodAbbr offers significant improvements over \LMPTAbbr in nearest-neighbor FT AUROC across all graph types with strong topic local consistency.

\subsubsection*{Semi-synthetic Result 2: \methodAbbr's empirical results are in agreement with theoretical findings}
In line with our theoretical findings, \methodAbbr only under-performs \LMPTAbbr on the \textsc{Structural} graph where the topic task (by design) does not have strong local consistency. This validates our theoretical results by showing that local consistency strongly predicts \notzeroshot FT performance.

\subsubsection*{Semi-synthetic Result 3: \methodAbbr is robust to incomplete and noisy pre-training graphs}
Figure~\ref{fig:table-noise-emb}b shows \notzeroshot FT AUROC as a function of noise rate on the \textsc{Cliques} graph. For up to 15\% noise, \methodAbbr shows improvements over \LMPTAbbr, and even at 50\% noise, the two approaches perform comparably.

\subsubsection*{Semi-synthetic Result 4: \methodAbbr pre-trained embeddings show stronger clustering than \LMPTAbbr embeddings}
Figure~\ref{fig:table-noise-emb}c-d shows embeddings produced under the \textsc{M\"obius} graph either by \LMPTAbbr or \methodAbbr, clustered via UMAP into 2 dimensions. It is clear visually from these figures that \methodAbbr embeddings show clear clusters strongly associated with the topic-modelling FT task, whereas \LMPTAbbr embeddings do not.

\subsubsection*{Conclusions}
From these analyses, we see that augmenting PT with per-sample structure-inducing objectives can both (1) offer significant advantages over existing PT architectures and (2) permit analytical reasoning about which FT tasks PT will offer improvements. 
These findings are not surprising; in these semi-synthetic experiments, we designed our graphs explicitly to have either high or low local consistency with respect to our FT task so that we could probe exactly whether SIPT methods would behave in accordance with theory in tightly controlled settings. In this way, the graphs $G_\PT$ used here may not be reflective of graphs in the real world, which will be chosen more independently of specific FT tasks. To address this, in the Results section, we demonstrate experimental results over diverse real-world datasets with real, FT-task-independent graphs to show that the gains persist in more realistic scenarios.

\subsection*{Further Details on Real-world Experiments}\label{methods_sec:further_details_on_experiments}

\subsubsection*{Further Details on the \textsc{Proteins} Dataset and FT tasks}
\begin{description}
    \item[PT Dataset] We use a dataset of $\sim$1.5M protein sequences from the Stanford Tree-of-life dataset~\cite{zitnik_evolution_2019} (\url{https://snap.stanford.edu/tree-of-life/data.html}). The associated Github repository for this resource lists an MIT license.
    
    \item[PT Graph] Two proteins are linked in $G_\PT$ if and only if they are documented in the scientific literature to interact, according to the tree-of-life interaction dataset. This is an external knowledge graph.
    
    \item[FT Dataset/Tasks] We use the TAPE FT benchmark tasks~\cite{rao_evaluating_2019}, including Remote homology (RH), a per-sequence classification task to predict protein fold category (metric: accuracy); Secondary structure (SS), a per-token classification task to predict amino acid structural properties (metric: accuracy); Stability (ST) \& Fluorescence (FL), per-sequence, regression tasks to predict a protein's stability and fluorescence, respectively (metric: Spearman's $\rho$); and Contact prediction (CP), an intra-sequence classification task to predict which pairs of amino acids are in contact in the protein's 3D conformation (metric: Precision at $L/5$). 
    
    \item[Baselines] We compare against the published TAPE model~\cite{rao_evaluating_2019}, which uses an LM task alone as our per-token comparison point, and the PLUS~\cite{PLUS_min_pre-training_2020} model, which optimizes for LM and supervised classification jointly, for our per-sample comparison point.
\end{description}

The tasks in the TAPE benchmark~\cite{rao_evaluating_2019} on which we test are described more fully below. All these datasets are publicly available.  All datasets can be obtained directly on TAPE's Github (\url{https://github.com/songlab-cal/tape#data}), which lists no licenses for these datasets though the overall Github is released under a BSD 3-Clause "New" or "Revised" License.
\begin{description}
    \item[Remote Homology] This is a per-sequence, multi-class classification problem, evaluated using accuracy, which tasks a model to predict a protein fold category at a per-sequence level. This task's dataset contains 12,312/736/718 train/val/test proteins and is originally sourced from~\cite{hou_deepsf_2018}.
    \item[Secondary Structure] This is a per-token, multi-class classification problem, evaluated using accuracy, which tasks a model to predict the structural properties of each amino acid in the final, folded protein. This task's dataset contains 8,678/2,170/513 train/val/test proteins, and is originally sourced from~\cite{klausen_netsurfp-20_2019}.
    \item[Stability] This is a per-sequence, continuous regression problem evaluated using the Spearman correlation coefficient, which tasks a model to predict the protein's stability in response to environmental conditions. This task's dataset contains 53,679/2,447/12,839 train/val/test proteins, and is originally sourced from~\cite{rocklin_global_2017}.
    \item[Fluorescence] This is a per-sequence, continuous regression problem evaluated using the Spearman correlation coefficient, which tasks a model to predict how brightly a protein will fluoresce. This task's dataset contains 21,446/5,362/27,217 train/val/test proteins, and is originally sourced from~\cite{sarkisyan_local_2016}.
\end{description}

\subsubsection*{Further Details on the \textsc{Abstracts} Dataset and FT tasks}

\begin{description}
    \item[PT Dataset] We use a dataset of $\sim$650K free-text scientific article abstracts from the Microsoft Academic Graph (MAG) dataset~\cite{wang_review_2019,hu_open_2020}. The \textsc{Abstracts} PT data (the Microsoft Academic Graph dataset) is licensed with an Open Data Commons Attribution License (ODC-By) v1.0 license.
    
    \item[PT Graph] Two abstracts are linked in $G_\PT$ if and only if their corresponding papers cite one another. This is a self-supervised graph.
    
    \item[FT Dataset/Task] We use a subset of the fine-tuning tasks used in the SciBERT paper~\cite{beltagy_scibert_2019}, including Paper field (PF), SciCite (SC), ACL-ARC (AA), and SciERC Relation Extraction (SRE), all of which are per-sentence classification problems (metric: Macro-F1). PF tasks models to predict a paper's area of study from its title, SC \& AA tasks both predict an ``intent'' label for citations, and SRE is a relation extraction task.
    
    \item[Baseline] We compare against the published SciBERT model~\cite{beltagy_scibert_2019} as our per-token comparison and lack an associated per-sample comparison as we don't know of any published per-sample models in the academic papers modality.
\end{description}

The tasks in the SciBERT benchmark~\cite{beltagy_scibert_2019} on which we test are described more fully below. All tasks here are per-sentence, multi-class classification problems (i.e., we do not study any per-token tasks), and all are evaluated in Macro-F1 (out of 1). All FT datasets can be obtained from the SciBERT Github (\url{https://github.com/allenai/scibert}), which lists no dataset-specific licenses but is released with an Apache-2.0 license.
\begin{description}
    \item[Paper Field] This problem asks models to predict a paper's area of study given its title. This task's dataset contains 84,000/5,599/22,399 train/val/test sentences. Though the original dataset is derived from the MAG~\cite{wang_review_2019}, it was formulated into this task format by SciBERT directly~\cite{beltagy_scibert_2019}.
    \item[SciCite] This problem tasks models to predict an ``intent'' label for sentences that cite other scientific works within academic articles. This task's dataset contains 7,320/916/1,861 train/val/test sentences, and is originally sourced from~\cite{cohan_structural_2019}.
    \item[ACL-ARC] This problem tasks models to predict an ``intent'' label for sentences that cite other scientific works within academic articles. This task's dataset contains 1,688/114/139 train/val/test sentences and is originally sourced from~\cite{jurgens_measuring_2018}.
\end{description}

\subsubsection*{Further Details on the \textsc{Networks} Dataset and FT tasks} 
\begin{description}
    \item[PT Dataset] We use a dataset of $\sim$70K protein-protein interaction (PPI) ego-networks here, sourced from~\cite{GRAPH_PT_hu_strategies_2019}. Each individual sample here describes a single protein, realized as a biological network (\ie, an attributed graph) corresponding to the ego-network about that protein (\ie, a small subgraph containing all nodes within the target protein) in a broader PPI graph. Unlike our other domains, this domain does not contain sequences. The \textsc{Networks} PT dataset releases its code and dataset files under an MIT license. 
    
    \item[PT Graph] The dataset from~\cite{GRAPH_PT_hu_strategies_2019} is labeled with the presence or absence of any of 4000  
    protein gene ontology terms associated with the central protein in each PPI ego network. Leveraging these labels, two PPI ego-networks are linked in $G_\PT$ if and only if the Hamming distance between their observed label vectors is no more than $9$. This is an alternate-representation nearest-neighbor graph.
    
    \item[FT Dataset/Tasks] Our FT task is the multi-label binary classification of the 40 gene-ontology term annotations (metric: macro-AUROC) used in~\cite{GRAPH_PT_hu_strategies_2019}. We use the PT set for FT training and evaluate the model on a held-out random 10\% split.
    
    \item[Baselines] We compare against both attribute-masking~\cite{GRAPH_PT_hu_strategies_2019} and multi-task supervised PT.
\end{description}

The Networks FT task is a multi-task, binary classification task. Recall that the dataset here consists of PPI ego-networks, which means that an individual sample input to the model is an attributed graph $\vec x$ which contains a central node, corresponding to a protein, along with the ego-graph surrounding that node in a larger PPI graph. This ego-graph can thus be seen to correspond to the central protein, and the FT and PT tasks leverage this association, as both of which flag whether or not that central protein is associated with particular gene-ontology (GO) terms (annotations relating to protein properties or function applied in the literature). The PT tasks contain 4000 possible GO annotations, but the FT tasks correspond to a smaller set of only 40 GO terms, chosen as they were of greater interest than the full set. See the original source (\cite{GRAPH_PT_hu_strategies_2019}) for more information and full details.

\subsubsection*{Further Details on Experimental Procedure}
To minimize computational burden, we do not pre-train a structure-inducing model from scratch for \textsc{Proteins} and \textsc{Abstracts} datasets. Instead, we initialize a model from the per-token baseline directly, then perform additional pre-training for only a small number of epochs under the new SIPT loss subdivision. We assess both multi-similarity and contrastive $\lossGGML$ variants in these domains. On the \textsc{Networks} dataset, we pre-train all models (including baselines) from scratch, and based on early experimental results, we only assess the contrastive loss variant.

\subsubsection*{Further Details on Ablation Studies}
Note that the warm-start procedure described above on the \textsc{Proteins} and \textsc{Abstracts} domains allows a powerful ablation study: by additionally training a PT model from the per-token baseline with $\lambdaGGML = 0$, we can uniquely assess the impact of the new loss term, rather than simply additional training or the different PT dataset. We perform this ablation study for all applicable datasets. For the \textsc{Networks} dataset, no additional ablation studies are needed to assess the impact of the loss term, given all models are trained from scratch with the same early-stop procedures.

\subsubsection*{Further Details on Choosing $\lambdaGGML$}
For the \textsc{Proteins} and \textsc{Abstracts} dataset, to choose the optimal value of $\lambdaGGML$ for use at PT time, we pre-trained several models and evaluated their efficacy in a link retrieval task on $G_\PT = (V, E)$. In particular, we score a node embedder $f$ by embedding all nodes $n \in V$ as $f(n)$, then rank all other nodes $n'$ by the euclidean distance between $f(n)$ and $f(n')$, and assess this ranked list via IR metrics including label ranking average precision (LRAP), normalized discounted cumulative gain (nDCG), average precision (AP), and mean reciprocal rank (MRR), where a node $n'$ is deemed to be a ``successful'' retrieval for $n$ if $(n, n') \in E$. In this way, note that we choose $\lambdaGGML$ in a manner that is independent of the fine-tuning task and can be determined solely based on the PT data. Final results for these experiments are shown in Methods Table~\ref{tab:pro_IR_results} for the proteins dataset and Methods Table~\ref{tab:sci_IR_results} for scientific articles.

Ultimately, this process suggests that $\lambdaGGML$ of $0.1$ is a robust setting, and as such, $0.1$ was used directly for the \textsc{Networks} task without further optimization.

\subsubsection*{Further Details on Architecture \& Hyperparameters}
The architectures of our encoders for the \textsc{Proteins} and \textsc{Abstracts} domains are fully determined from our source models in TAPE~\cite{rao_evaluating_2019} and SciBERT~\cite{beltagy_scibert_2019}. In particular, for proteins and scientific articles, we use a 12-layer Transformer with a hidden size of 768, an intermediate size of 3072, and 12 attention heads. Provided TAPE and SciBERT tokenizers are also used. A single linear layer to the output dimensionality of each task is used s the prediction head, taking as input the output of the final layer's \texttt{[CLS]} token as a whole-sequence embedding. We also tested either pre-training for a single or for four additional epochs, based on validation set performance, and ultimately used a single epoch for proteins and four for scientific articles. 

For the \textsc{Networks} domain, we match the architecture used in the original source~\cite{GRAPH_PT_hu_strategies_2019} for the mask model runs. Save that for computational efficiency, we scale the batch size up as high as it can go, then proportionally scale up the learning rate to account for the larger batch size. This corresponds to a batch size of 1024, the learning rate of 0.01, a GCNN encoder type of GIN, embedding dimensions of 300, 5 layers, 10\% dropout, mean pooling, and a JK strategy of ``last''.

Fine-tuning hyperparameters (learning rate, batch size, and the number of epochs) were determined based on a combination of existing results, hyperparameter tuning, and machine limitations. On proteins, most hyperparameters were set to follow those reported for a \LMPTAbbr model in \cite{mcdermott2021adversarial}, though additional limited hyperparameter searches were performed to validate that these choices were adequate. As the original source for these hyperparameters was an \LMPTAbbr model, any bias here should be \emph{against} \methodAbbr, meaning this is a conservative choice. Early stopping (based on the number of epochs without observing improvement in the validation set performance) was employed, and batch size was set as large as possible given the limitations of the underlying machine. For the PLUS reproduction, we compared hyperparameters analogous to the reported PLUS hyperparameters for other tasks and analogous to our hyperparameters for other tasks and used those that performed best on the validation set. For scientific articles, we performed a grid search to optimize downstream task performance on the validation set, with the learning rate varying between 5e-6 and 5e-5 and the number of epochs between 2 and 5. The same grid search was used in the original SciBERT method. We additionally match the SciBERT benchmark by applying a dropout of 0.1, using the Adam optimizer with linear warm-up and decay, a batch size of 32, and no early stopping. For the \textsc{Networks}, FT hyperparameters were again chosen to match the original source model~\cite{GRAPH_PT_hu_strategies_2019} to save the increase in batch size and learning rate. No additional hyperparameter search was performed.

Final hyperparameters for each downstream task are shown in Tables~\ref{tab:tape_hyperparams} for proteins and \ref{tab:sci_hyperparameters} for scientific articles.

\begin{table}
    \centering
    \caption{Final hyperparameters for our \textsc{Proteins} domain. All tasks used 200 total epochs and performed early stopping after 25 epochs of no validation set improvement. LR, learning rate. \\}
    \begin{tabular}{lrr} \toprule
        Task                & Batch Size & LR   \\ \midrule
        Remote Homology     & 16         & 1e-5 \\
        Fluorescence        & 128        & 5e-5 \\
        Stability           & 512        & 1e-4 \\
        Secondary Structure & 16         & 1e-5 \\
    \bottomrule \end{tabular}
    \label{tab:tape_hyperparams}
\end{table}

\begin{table}
    \centering
    \caption{Final hyperparameters for our \textsc{Abstracts} dataset. All models used a batch size of 32 and no early stopping to match the original SciBERT paper~\cite{beltagy_scibert_2019}. LR, learning rate. A / B = [\LMPTAbbr Hyperparameter] / [\methodAbbr Hyperparameter]. \\}
    \begin{tabular}{lrr} \toprule
        Task         & \# Epochs & LR   \\ \midrule
        Paper Field  & 2         & 5e-5 \\
        ACL-ARC      & 4/5       & 5e-5 \\
        SciCite      & 3/2       & 1e-5 \\
    \bottomrule \end{tabular}
    \label{tab:sci_hyperparameters}
\end{table}

\subsubsection*{Further Details on Implementation and Compute Environment}
We leverage PyTorch for our codebase. FT Experiments and \textsc{Networks} PT were run over various ubuntu machines (versions ranged from 16.04 to 20.04) with a variety of NVIDIA GPUs. \textsc{Proteins} and \textsc{Abstracts} PT runs were performed on a Power 9 system, each run using 4 NVIDIA 32 GB V100 GPUs with InfiniBand at half precision.

\subsubsection*{Full Results} \label{appendix:subsec:raw_results}
Here we provide the raw FT results for all tasks in the \textsc{Proteins} and \textsc{Abstracts} domains, respectively (Tables~\ref{tab:tape_FT_results}~and~\ref{tab:sciBERT_FT_results}). The \textsc{Networks} domain raw results are already present in the main text (Figure~\ref{fig:graph_pt_res}).

\begin{table}
    \centering
    \begin{tabular}{p{0.12\linewidth}rrrrr} \toprule
        Model         & RH                       & FL                      & ST                      & SS                       & CP            \\ \midrule
        TAPE          & 21\%                     & \textbf{0.68}           & 0.73                    & 73\%                     & 0.32          \\ 
        PLUS          & 19.8\%\std{1.7}$^*$      & 0.63                    & 0.76                    & 73\%                     & N/A           \\
        \LMPTAbbr     & 23.8\%\std{1.1}          & 0.67\std{0.00}          & 0.76\std{0.02}          & 73.9\%\std{0.0}          & 0.38          \\ \midrule
        \methodAbbr-C & 25.1\%\std{0.6}          & \textbf{0.68\std{0.00}} & \textbf{0.77\std{0.01}} & 73.9\%\std{0.0}          & 0.38          \\
        \methodAbbr-M & \textbf{26.6\%\std{1.0}} & \textbf{0.68\std{0.00}} & 0.76\std{0.01}          & \textbf{74.2\%\std{0.1}} & \textbf{0.39} \\
    \bottomrule \end{tabular}
    \caption{
        Results of the TAPE Transformer~\cite{rao_evaluating_2019}, the PLUS Transformer~\cite{PLUS_min_pre-training_2020} ($^*$: our measurements), our \LMPTAbbr baseline, and two \methodAbbr variants (``-C'' indicates the contrastive loss, ``-M'' the multisimilarity loss). Higher is better. 
    }
    \label{tab:tape_FT_results}
\end{table}

\begin{table}
    \centering
    \begin{tabular}{p{0.12\linewidth}rrrr} \toprule
        Model         & PF                     & SC                      & AA                      & SRE                     \\ \midrule
        SciBERT       & \textbf{0.66}          & 0.85                    & 0.71                    & 0.80                    \\ 
        \LMPTAbbr     & \textbf{0.66\std{0.0}} & 0.85\std{0.01}          & 0.70\std{0.05}          & 0.80\std{0.01}          \\ \midrule
        \methodAbbr-C & \textbf{0.66\std{0.0}} & \textbf{0.86\std{0.01}} & \textbf{0.76\std{0.02}} & \textbf{0.81\std{0.00}} \\
        \methodAbbr-M & \textbf{0.66\std{0.0}} & 0.85\std{0.00}          & 0.73\std{0.05}          & N/A                     \\
    \bottomrule \end{tabular}
    \caption{
        Results of the original SciBERT~\cite{beltagy_scibert_2019} model, our own \LMPTAbbr baseline, and two \methodAbbr variants (``-C'' indicates the contrastive loss, ``-M'' the multisimilarity loss). Higher is better.
    }
    \label{tab:sciBERT_FT_results}
\end{table}

\subsubsection*{SIPT Results are in Accordance with Theory and Guiding Hypothesis} \label{appendix:subsec:real_data_theory}

Results over all real-world domains are consistent with our theoretical analyses and guiding hypothesis. We can also analyze the extent to which induced structure helps non-NLP domains by examining the results of our $\lambdaGGML$ tuning procedure.
In particular, we find that far less structure-inducing is necessary on our \textsc{Abstracts} dataset ($\lambdaGGML=0.01$) than on our \textsc{Proteins} dataset ($\lambdaGGML=0.1$). This agrees with our guiding hypothesis that per-sample latent space regularization is much more necessary on non-NLP domains than on NLP domains. 

To demonstrate this, we show the final results for the guiding link-retrieval task for the \textsc{Proteins} domain in Table~\ref{tab:pro_IR_results} and for the \textsc{Abstracts} domain in Table~\ref{tab:sci_IR_results}. In both settings, we compare the following models.
\begin{description}
    \item[Random] Nodes are embedded with random vectors to assess chance performance.
    \item[Initial Model] Nodes are embedded with the base pre-trained model we build on in our experiments without further modifications. This model is TAPE~\cite{rao_evaluating_2019} for proteins and SciBERT~\cite{beltagy_scibert_2019} for scientific articles.
    \item[\LMPTAbbr] Nodes are embedded with the final encoder after additional pre-training on our graph-augmented datasets, but without any \methodAbbr (i.e., $\lambdaGGML = 0$).
    \item[CS RoBERTa] \textit{(for scientific articles only)} Nodes are embedded via \cite{DAPT_gururangan-etal-2020-dont}'s DAPT CS RoBERTa model, which is another \LMPTAbbr model over scientific abstracts which performed very well on ACL-ARC, the task on which \methodAbbr does best in scientific articles.
    \item[\methodAbbr] \textit{(for various values of $\lambdaGGML$)}. Nodes are represented via \methodAbbr PT models at the specified weighting. For proteins, all \methodAbbr models are initialized from TAPE, but for scientific articles, we test against both initializing from SciBERT and CS RoBERTa (as both are just different, domain-specific \LMPTAbbr models).
\end{description}

Note that in addition to the discrepancy in the magnitude of improvement (over scientific articles, average precision goes from 12.9\% to 14.2\%, vs. 2.4\% to 3.5\% on proteins, which is proportionally much more significant), we can also see that \methodAbbr improves retrieval performance over the baselines for proteins much more than it does for scientific articles. This is, admittedly, largely due to \cite{DAPT_gururangan-etal-2020-dont}'s CS RoBERTa model's surprisingly good performance without any modifications, however as we also compare \methodAbbr pre-trained from a CS RoBERTa model and it does not demonstrate significant improvements, we still feel this is a fair comparison. These findings are consistent with our hypothesis that \methodAbbr will offer more significant advantages in non-natural language domains.

\begin{table*}
    \centering
    \caption{PT set link-retrieval performance for a random baseline, the raw TAPE model, and \methodAbbr for various weighting parameters $\lambdaGGML$ on the dataset of protein sequences. LRAP, label ranking average precision; nDCG, normalized discounted cumulative gain; AP, average precision; MRR, mean reciprocal rank. Higher values indicate better performance. Highlighted in grey are realizations of \methodAbbr framework that yield better results than the strongest baseline, providing evidence that incorporating sequence-level relational information into PT (\ie, $\lambdaGGML > 0$) leads to improved performance.}
    \begin{tabular}{lrrrrr} \toprule
        Method          & $\lambdaGGML$ & LRAP                       & nDCG                      & AP                        & MRR                      \\ \midrule
        Random Baseline & N/A           & 0.88\%                     & 27.1\%                    & 0.88\%                    & 0.003                    \\ 
        TAPE~\cite{rao_evaluating_2019}
                        & N/A           & 8.50\%                     & 34.9\%                    & 2.41\%                    & 0.226                    \\ 
        \LMPTAbbr Baseline
                        & 0             & 8.92\%                     & 38.0\%                    & 2.33\%                    & 0.238                    \\ \midrule
        \multirow{5}{*}{\methodAbbr (TAPE Initialized)}
                        & 0.01          & \cellcolor{gray!25}9.69\%  & \cellcolor{gray!25}39.1\% & \cellcolor{gray!25}2.56\% & \cellcolor{gray!25}0.254 \\
                        & 0.10          & \cellcolor{gray!25}10.95\% & \cellcolor{gray!25}39.4\% & \cellcolor{gray!25}3.46\% & \cellcolor{gray!25}0.260 \\
                        & 0.50          & \cellcolor{gray!25}10.54\% & \cellcolor{gray!25}40.3\% & \cellcolor{gray!25}3.43\% & \cellcolor{gray!25}0.246 \\
                        & 0.90          & \cellcolor{gray!25}10.12\% & \cellcolor{gray!25}39.0\% & \cellcolor{gray!25}3.16\% & 0.237                    \\
                        & 0.99          & \cellcolor{gray!25}14.50\% & 37.5\%                    & \cellcolor{gray!25}3.13\% & 0.236                    \\
    \bottomrule \end{tabular}
    \label{tab:pro_IR_results}
\end{table*}

\begin{table*}
    \centering
    \caption{PT set link-retrieval performance for a random baseline, the raw SciBERT model, and \methodAbbr for various weighting parameters $\lambdaGGML$ on the scientific articles dataset. LRAP, label ranking average precision; nDCG, normalized discounted cumulative gain; AP, average precision; MRR, mean reciprocal rank. Higher values indicate better performance. Highlighted in grey are realizations of \methodAbbr framework that yield better results than the strongest baseline, providing evidence that incorporating sequence-level relational information into PT (\ie, $\lambdaGGML > 0$) leads to improved performance.}
    \begin{tabular}{lrrrrr} \toprule
        Method          & $\lambdaGGML$ & LRAP                       & nDCG                      & AP                         & MRR                      \\\midrule
        Random Baseline & N/A           & 0.89\%                     & 26.0\%                    & 0.27\%                     & 0.016                    \\ 
        SciBERT~\cite{beltagy_scibert_2019}
                        & N/A           & 17.22\%                    & 52.8\%                    & 5.16\%                     & 0.272                    \\
        \LMPTAbbr Baseline (SciBERT initialized)
                        & 0             & 16.79\%                    & 35.4\%                    & 5.00\%                        & 0.271                    \\
        DAPT CS RoBERTa~\cite{DAPT_gururangan-etal-2020-dont}
                        & N/A           & 32.56\%                    & 50.3\%                    & 12.86\%                    & 0.459                    \\
        \LMPTAbbr Baseline (CS RoBERTa initialized)
                        & 0             & 30.58\%                    & 48.3\%                    & 12.36\%                    & 0.438                    \\ \midrule
        \multirow{5}{*}{\methodAbbr (SciBERT initialized)}    
                        & 0.01          & \cellcolor{gray!25}42.26\% & \cellcolor{gray!25}58.7\% & \cellcolor{gray!25}14.23\% & \cellcolor{gray!25}0.536 \\
                        & 0.10          & \cellcolor{gray!25}34.73\% & 52.5\%                    & 9.39\%                     & 0.457                    \\
                        & 0.50          & \cellcolor{gray!25}32.85\% & 50.8\%                    & 8.37\%                     & 0.438                    \\
                        & 0.90          & 31.61\%                    & 49.8\%                    & 7.82\%                     & 0.426                    \\
                        & 0.99          & 30.72\%                    & 49.0\%                    & 6.80\%                      & 0.415                    \\ \midrule
        \multirow{5}{*}{\methodAbbr (CS RoBERTa initialized)}    
                        & 0.01          & \cellcolor{gray!25}33.32\% & 51.2\%                    & 8.61\%                     & 0.448                    \\
                        & 0.10          & 25.46\%                    & 44.4\%                    & 5.88\%                     & 0.359                    \\
                        & 0.50          & 25.08\%                    & 44.0\%                    & 6.08\%                     & 0.355                    \\
                        & 0.90          & 22.43\%                    & 41.6\%                    & 4.27\%                     & 0.317                    \\
                        & 0.99          & 22.38\%                    & 41.5\%                    & 4.68\%                     & 0.316                    \\
    \bottomrule \end{tabular}
    \label{tab:sci_IR_results}
\end{table*}

%% file: 051pt_methods_table.tex
\newcommand{\chk}{\checkmark}
\newcommand{\rbox}[1]{\rotatebox{90}{#1}}
\renewcommand{\arraystretch}{0.7}
\begin{table}
    \centering \tiny
    \begin{tabular}{p{0.25\linewidth}cccccccccc|ggggggg}
        \multirow{2}{*}{Method}
            & \rbox{Masked, discriminative, or standard language modelling}
            & \rbox{Template/prompt-style multi-task language model training}
            & \rbox{Concatenate related sentences together}
            & \rbox{Named entity masking}
            & \rbox{Relation masking}
            & \rbox{Per-token knowledge graph alignment}
            & \rbox{Named entity recognition and linking}
            & \rbox{(Unconstrained) attention over a KG}
            & \rbox{Joint token and entity embeddings}
            & \rbox{Syntatic Knowledge Distillation}
            & \rbox{Single-task classification}
            & \rbox{Multi-task classification}
            & \rbox{Whole-sample graph alignment}
            & \rbox{Per-sample augmentation-based contrastive alignment}
            & \rbox{Multi-lingual cross-sample contrastive alignment}
            & \rbox{Unsupervised clustering}
            & \rbox{Contextual autoencoding}
        \\ \cmidrule(lr{5pt}){2-11} \cmidrule(lr{5pt}){12-18}
            & \multicolumn{10}{c}{\textbf{Per-token}}
            & \multicolumn{7}{|c}{\textbf{Per-sample}}
        \\ \midrule
        \cite{ELMO_peters_deep_2018}~ELMO
            & \chk &      &      &      &      &      &      &      &      &      &      &      &      &      &      &      &      \\
        \cite{GPTIII_brown2020language}~GPT-3
            & \chk &      &      &      &      &      &      &      &      &      &      &      &      &      &      &      &      \\
        \cite{T5_JMLR:v21:20-074}~T5
            &      & \chk &      &      &      &      &      &      &      &      &      &      &      &      &      &      &      \\
        \cite{liu_roberta_2019}~RoBERTa
            & \chk &      &      &      &      &      &      &      &      &      &      &      &      &      &      &      &      \\
        \cite{GPTI_radford2018improving}~GPT-1
            & \chk &      &      &      &      &      &      &      &      &      &      &      &      &      &      &      &      \\
        \cite{GPTII_radford2019language}~GPT-2
            & \chk &      &      &      &      &      &      &      &      &      &      &      &      &      &      &      &      \\
        \cite{BART_lewis-etal-2020-bart}~BART
            & \chk &      &      &      &      &      &      &      &      &      &      &      &      &      &      &      &      \\
        \cite{UNSUP_XLNG_conneau-etal-2020-unsupervised}~Unsupervised Cross Lingual
            & \chk &      &      &      &      &      &      &      &      &      &      &      &      &      &      &      &      \\
        \cite{ELECTRA_clark_electra_2019}~ELECTRA
            & \chk &      &      &      &      &      &      &      &      &      &      &      &      &      &      &      &      \\
        \cite{SPANBERT_joshi-etal-2020-spanbert}~SpanBERT
            & \chk &      &      &      &      &      &      &      &      &      &      &      &      &      &      &      &      \\
        \cite{UNI_LM_NEURIPS2019_c20bb2d9}~UniLM
            & \chk &      &      &      &      &      &      &      &      &      &      &      &      &      &      &      &      \\
        \cite{DAPT_gururangan-etal-2020-dont}~DAPT
            & \chk &      &      &      &      &      &      &      &      &      &      &      &      &      &      &      &      \\
        \cite{ERNIE1_sun_ernie_2019}~ERNIE (Sun et. al.)
            & \chk &      &      & \chk &      &      &      &      &      &      &      &      &      &      &      &      &      \\
        \cite{KNOWBERT_peters2019knowledge}~KnowBERT
            & \chk &      &      &      &      &      & \chk & \chk & \chk &      &      &      &      &      &      &      &      \\
        \cite{rao_evaluating_2019}~TAPE
            & \chk &      &      &      &      &      &      &      &      &      &      &      &      &      &      &      &      \\
        \cite{LUKEyamada2020luke}~LUKE
            & \chk &      &      & \chk &      &      &      &      &      &      &      &      &      &      &      &      &      \\
        \cite{T0PP_sanh2022multitask}~T0pp
            &      & \chk &      &      &      &      &      &      &      &      &      &      &      &      &      &      &      \\
        \cite{PRE_ENC_Xiong2020Pretrained}~Pretrained Encyclopedia
            & \chk &      &      & \chk &      &      &      &      &      &      &      &      &      &      &      &      &      \\
        \cite{MSA_pmlr-v139-rao21a}~MSA
            & \chk &      & \chk &      &      &      &      &      &      &      &      &      &      &      &      &      &      \\
        \cite{COLAKE_sun-etal-2020-colake}~COLAKE
            & \chk &      &      & \chk & \chk &      &      &      &      &      &      &      &      &      &      &      &      \\
        \cite{BERTMK_he-etal-2020-bert}~BERTMK
            & \chk &      &      &      &      &      &      &      & \chk &      &      &      &      &      &      &      &      \\
        \cite{ERICA_qin2020erica}~ERICA
            & \chk &      &      &      &      & \chk &      &      &      &      &      &      &      &      &      &      &      \\
        \cite{JAKET_yu2020jaket}~JAKET
            & \chk &      &      &      &      &      &      &      & \chk &      &      &      &      &      &      &      &      \\
        \cite{CALM_zhou2021pretraining}~CALM
            &      & \chk &      &      &      &      &      &      &      &      &      &      &      &      &      &      &      \\
        \cite{KEBIOLM_yuan-etal-2021-improving}~KeBioLM
            & \chk &      &      &      &      &      & \chk & \chk & \chk &      &      &      &      &      &      &      &      \\
        \cite{MGBERT_zhang2021mg}~MG-BERT (Molecules)
            & \chk &      &      &      &      &      &      &      &      &      &      &      &      &      &      &      &      \\
        \cite{CDLM_caciularu-etal-2021-cdlm-cross}~CDLM
            & \chk &      & \chk &      &      &      &      &      &      &      &      &      &      &      &      &      &      \\
        \cite{KGPLM_he2020kgplm}~KgPLM
            & \chk &      &      & \chk &      &      &      &      &      &      &      &      &      &      &      &      &      \\
        \cite{KNN_PT_levine2022the}~kNN PT
            & \chk &      & \chk &      &      &      &      &      &      &      &      &      &      &      &      &      &      \\
        \cite{LP-BERT_DBLP:journals/corr/abs-2201-04843}~LP-BERT
            & \chk &      &      & \chk & \chk &      &      &      &      &      &      &      &      &      &      &      &      \\
        \cite{MGBERT_behnamghader2021mg}~MG-BERT (NLP)
            & \chk &      &      &      &      &      &      &      & \chk &      &      &      &      &      &      &      &      \\
        \cite{UD-PRLM_NEURIPS2021_473803f0}~UD-PrLM
            & \chk &      &      &      &      & \chk &      &      &      &      &      &      &      &      &      &      &      \\
        \cite{rives_biological_2019}~ESM-1B
            & \chk &      &      &      &      &      &      &      &      &      &      &      &      &      &      &      &      \\
        \cite{alley_unified_2019}~UniRep
            & \chk &      &      &      &      &      &      &      &      &      &      &      &      &      &      &      &      \\
        \cite{devlin_bert_2019}~BERT
            & \chk &      &      &      &      &      &      &      &      &      & \chk &      &      &      &      &      &      \\
        \cite{ERNIE_ALT_zhang_ernie_2019}~ERNIE (Zhang et. al.)
            & \chk &      &      & \chk &      &      &      &      & \chk &      & \chk &      &      &      &      &      &      \\
        \cite{COKEBERT_su2021cokebert}~CokeBERT
            & \chk &      &      & \chk &      &      &      &      & \chk &      & \chk &      &      &      &      &      &      \\
        \cite{SPIDER_zhang-zhao-2021-structural}~SPIDER
            & \chk &      &      &      &      & \chk &      &      &      &      & \chk &      &      &      &      &      &      \\
        \cite{kuncoro-etal-2020-syntactic}~Syntatic-Distilled BERT
            & \chk &      &      &      &      &      &      &      &      & \chk & \chk &      &      &      &      &      &      \\
        \cite{lan_albert_2019}~ALBERT
            & \chk &      &      &      &      &      &      &      &      &      & \chk &      &      &      &      &      &      \\
        \cite{SMEDBERT_zhang2021smedbert}~SMedBERT
            & \chk &      &      &      &      & \chk &      &      & \chk &      & \chk &      &      &      &      &      &      \\
        \cite{MT_DNN_liu_multi-task_2019}~MT-DNN
            & \chk &      &      &      &      &      &      &      &      &      &      & \chk &      &      &      &      &      \\
        \cite{GRAPH_PT_hu_strategies_2019}~Graph-PT
            & \chk &      &      &      &      &      &      &      &      &      &      & \chk &      &      &      &      &      \\
        \cite{SENTILARE_ke-etal-2020-sentilare}~SentiLARE
            & \chk &      &      &      &      &      &      &      &      &      & \chk &      &      &      &      &      &      \\
        \cite{PLUS_min_pre-training_2020}~PLUS
            & \chk &      &      &      &      &      & \chk &      &      &      & \chk &      &      &      &      &      &      \\
        \cite{mcdermott_comprehensive_2020}~EHR-PT
            & \chk &      &      &      &      &      &      &      &      &      &      & \chk &      &      &      &      &      \\
        \cite{ERNIE2_sun_ernie_2020}~ERNIE 2.0 (Sun et. al.)
            & \chk &      &      & \chk &      &      &      &      &      &      &      & \chk &      &      &      &      &      \\
        \cite{ERNIE3_sun2021ernie}~ERNIE 3.0 (Sun et. al.)
            & \chk &      &      & \chk & \chk &      &      &      &      &      &      & \chk &      &      &      &      &      \\
        \cite{Dict-BERT-yu-etal-2022-dict}~Dict-BERT
            & \chk &      & \chk &      &      &      & \chk &      &      &      & \chk &      &      &      &      &      &      \\
        \cite{LINKBERT_yasunaga-etal-2022-linkbert}~LinkBERT
            & \chk &      &      &      &      &      &      &      &      &      & \chk &      &      &      &      &      &      \\
        \cite{STRUCTBERT_Wang2020StructBERT}~StructBERT
            & \chk &      &      &      &      &      &      &      &      &      & \chk &      &      &      &      &      &      \\
        \cite{MARGE_lewis2020pre}~MARGE
            &      &      &      &      &      &      &      &      &      &      &      &      &      &      &      & \chk & \chk \\
        \cite{REALM_10.5555/3524938.3525306}~REALM
            & \chk &      & \chk & \chk &      &      &      &      &      &      &      &      &      &      &      & \chk &      \\
        \cite{GraphCL_NEURIPS2020_3fe23034}~GraphCL
            &      &      &      &      &      &      &      &      &      &      &      &      &      & \chk &      &      &      \\
        \cite{GCC_10.1145/3394486.3403168}~GCC
            &      &      &      &      &      &      &      &      &      &      &      &      &      & \chk &      &      &      \\
        \cite{DECLUTR_giorgi-etal-2021-declutr}~DeCLUTR
            & \chk &      &      &      &      &      &      &      &      &      &      &      &      & \chk &      &      &      \\
        \cite{CLEAR_wu2020clear}~CLEAR
            & \chk &      &      &      &      &      &      &      &      &      &      &      &      & \chk &      &      &      \\
        \cite{JOAO_pmlr-v139-you21a}~JOAO
            &      &      &      &      &      &      &      &      &      &      &      &      &      & \chk &      &      &      \\
        \cite{COCO_LM_NEURIPS2021_c2c2a045}~COCO-LM
            & \chk &      &      &      &      &      &      &      &      &      &      &      &      & \chk &      &      &      \\
        \cite{INFOWORD_Kong2020A}~InfoWord
            & \chk &      &      &      &      &      &      &      &      &      &      &      &      & \chk &      &      &      \\
        \cite{MICRO_GRAPH_zhang2020motif}~MICRO-Graph
            &      &      &      &      &      &      &      &      &      &      &      &      &      & \chk &      & \chk &      \\
        \cite{STS_CT_carlsson2021semantic}~STS-CT
            & \chk &      &      &      &      &      &      &      &      &      &      &      &      & \chk &      &      &      \\
        \cite{CAPT_luo2020capt}~CAPT
            &      &      &      &      &      &      &      &      &      &      &      &      &      & \chk &      &      &      \\
        \cite{GEARNET_zhang2022protein}~GearNet
            & \chk &      &      &      &      & \chk &      &      &      &      &      &      &      & \chk &      &      &      \\
        \cite{INFOXLM_chi-etal-2021-infoxlm}~InfoXLM
            & \chk &      &      &      &      &      &      &      &      &      &      &      &      &      & \chk &      &      \\
        \cite{GLM_shen-etal-2020-exploiting}~GLM
            & \chk &      &      & \chk &      &      &      &      &      &      &      &      &      & \chk &      &      &      \\
        \cite{KCL_fang2022molecular}~KCL
            &      &      &      &      &      &      &      &      &      &      &      &      &      & \chk &      &      &      \\
        \cite{wang_kepler_2019}~KEPLER
            & \chk &      &      &      &      &      &      &      &      &      &      &      & \chk &      &      &      &      \\
        \cite{CK_GNN_fang2021knowledge}~CK-GNN
            &      &      &      &      &      &      &      &      &      &      &      &      & \chk &      &      &      &      \\
        \cite{XLM_K_jiang2022xlm}~XLM-K
            & \chk &      &      &      &      &      &      &      &      &      &      &      & \chk &      &      &      &      \\
        \cite{WEBFORMER_guo2022webformer}~Webformer
            & \chk &      &      &      &      &      &      &      &      &      &      & \chk & \chk & \chk &      &      &      \\
    \bottomrule \end{tabular}
    \caption{
        \textbf{Existing Pre-training (PT) Methods:}
        A subset of existing PT methods, broken down by how they constrain per-token and per-sample latent space geometries.
    }
    \label{tab:existing_pt_methods}
\end{table}

%% file: 060appendix.tex
\section{Review of Language Model Pre-training Methods}
\label{sec:existing_LMPT_details}
In this supplementary section, we describe all of the models featured in our review (Figure~\ref{fig:existing_pt_methods} and Table~\ref{tab:existing_pt_methods}) and highlight key details of their approach. 

\subsection{Language modelling alone}
\begin{description}
    \item[\cite{ELMO_peters_deep_2018}] General domain NLP; ELMO leverages a biLSTM to perform language modelling; unlike later methods, for FT tasks, models do not typically re-train the entire LSTM but rather use a weighted combination of model interior hidden states as (at FT time) static word-embeddings.
    \item[\cite{liu_roberta_2019}] General domain NLP; RoBERTa includes only a masked language modelling objective.
    \item[\cite{GPTI_radford2018improving,GPTII_radford2019language,GPTIII_brown2020language}] General domain NLP; The GPT series of models use autoregressive language modelling alone and focus on generative language tasks, not general PT/FT, though GPT-III does show that by reframing many classical NLP fine-tuning tasks as generative language tasks, GPT-III can still offer a compelling zero and few-shot solution to these tasks using only the pre-trained embedder~\cite{GPTIII_brown2020language}.
    \item[\cite{BART_lewis-etal-2020-bart}] General domain NLP; BART utilizes a denoising language-model objective across various noising constraints.
    \item[\cite{UNI_LM_NEURIPS2019_c20bb2d9}] General domain NLP; UniLM integrates several different kinds of language modelling, including bidirectional, unidirectional, and sequence-to-sequence LMs. They impose no other PT losses.
    \item[\cite{rao_evaluating_2019,rives_biological_2019,alley_unified_2019}] Protein sequences; Various methods have explored language modelling alone for protein sequences. One notable entry is the TAPE benchmark, which also introduces a public benchmark of FT tasks for future comparisons.
    \item[\cite{MGBERT_zhang2021mg}] Molecular Graphs; Molecular Graph BERT (MG-BERT; no relation to MG-BERT~\cite{MGBERT_behnamghader2021mg}) uses masked atom prediction to pre-train a GNN over molecular graphs.
    \item[\cite{UNSUP_XLNG_conneau-etal-2020-unsupervised}] General domain NLP; This paper pre-trains a model for multi-lingual language modelling, using only a multi-lingual masked language modelling objective. 
    \item[\cite{DAPT_gururangan-etal-2020-dont}] General domain NLP; DAPT advocates for continual pre-training on increasingly task-focused text to improve its relevance to various downstream tasks. DAPT uses a RoBERTa baseline pre-training model, which includes only a masked language modelling objective. It shows significant gains after adaptation. However, as they only adapt the pre-training context to the more focused text, this induces no additional constraints on the latent space geometry.
    \item[\cite{SPANBERT_joshi-etal-2020-spanbert}] General domain NLP; SpanBERT changes the traditional masked language modelling task to a task in which contiguous spans are masked wholesale, rather than individual tokens.
\end{description}

\subsection{Language modelling \& templated tasks/prompting as language modelling}
\begin{description}
    \item[\cite{T5_JMLR:v21:20-074}] General domain NLP; T5 not only performs a robust analysis of various existing pre-training strategies but also introduces the ``text-to-text'' style of diverse pre-training, in which various downstream NLP tasks can be re-realized as language modelling tasks via templating and prompting, then integrated into language model pre-training alongside unsupervised objectives (such as traditional masked language modelling, albeit realized as a sequence-to-sequence task). As they realize all these downstream tasks as additional language modelling tasks, they neither officially produce a directly constrained per-sample embedding nor constrain the geometry of $\mathcal Z$ beyond traditional masked language modelling.
    \item[\cite{CALM_zhou2021pretraining}] General domain NLP; CALM builds on ideas from T5 to propose a text-to-text pre-training objective that leverages recognized per-token KG entities from the source text as a generative prompt.
    \item[\cite{T0PP_sanh2022multitask}] General domain NLP; T0pp extends the architecture of T5~\cite{T5_JMLR:v21:20-074} to ingest templated language modelling task from a wide variety of possible input tasks, then evaluates its performance in a zero-shot manner on unseen fine-tuning tasks.
\end{description}

\subsection{Language modelling \& Per-token KG Integration}
\begin{description}
    \item[\cite{ERNIE1_sun_ernie_2019}] General domain NLP; ERNIE 1 augments traditional MLM with entity-specific masking (e.g., masking the word ``Mozart'' from the sentence ``Mozart was a musician'') to force the model to recover common-sense knowledge about named entities.
    \item[\cite{KGPLM_he2020kgplm}] General domain NLP; KgPLM adapts the discriminative training ideas of ELECTRA~\cite{ELECTRA_clark_electra_2019} alongside the idea of entity masking explored previously. They perform entity masking and a discriminative loss identifying which tokens were replaced focused on entity replacements.
    \item[\cite{ERICA_qin2020erica}] General domain NLP; ERICA presents a mechanism for leveraging contrastive learning and distant supervision to incorporate external knowledge into a PLM for improving language understanding. ERICA augments MLM with two per-token tasks to ensure the per-token representations within a document reflect the structure of the KG. First, ERICA ensures that the pooled representations of head and tail entities are similar when conditioned on a relation (which is prepended to the document prior to embedding). Second, ERICA ensures that relation embeddings (defined as concatenated head, tail per-token entity embeddings) are similar within and across documents. As both tasks are done on per-token embeddings and never at a per-sample level, this approach induces minimal constraints on the per-sample latent space.
    \item[\cite{KNOWBERT_peters2019knowledge}] General domain NLP; Know-BERT integrates per-token entity information into an MLM pre-training scheme by performing unconstrained attention over a per-entity knowledge graph (only on pre-identified candidate entity spans), alongside any available entity linking supervision information via direct Named Entity Linking. This has similarities with \cite{KEBIOLM_yuan-etal-2021-improving} and \cite{ENTEXP_fevry-etal-2020-entities}.
    \item[\cite{KEBIOLM_yuan-etal-2021-improving}] Biomedical domain NLP; KeBioLM integrates a per-token KG into a biomedical language model by augmenting token entity representations with attention lookups into a biomedical KG (regardless of whether the attended entities match a given entity mention in the source text, though they do only apply this on recognized entities). To ensure this attention is meaningful, they perform named entity linking and recognition as auxiliary PT objectives, leveraging the same KG embeddings used during the attention calculation. In doing so, the method incentivizes per-token representations to be similar to their associated entity representations, thus ensuring that the entities are reflected in the attention over the KG. KG embeddings are initialized using Trans-E~\cite{TRANS_E_mroueh2012multiclass}. Their usage of automatically attending over entities within their language model (without explicit constraints on those matches) is motivated by~\cite{ENTEXP_fevry-etal-2020-entities}'s work in \cite{ENTEXP_fevry-etal-2020-entities} and has similarities to Know-BERT~\cite{KNOWBERT_peters2019knowledge}.
    \item[\cite{LUKEyamada2020luke}] General domain NLP; LUKE performs pre-training using MLM and an entity-specific masking/recognition scheme that is a slight variation on the traditional entity-specific masking \cite{ERNIE1_sun_ernie_2019} proposed. At FT time, they have other knowledge-specific integrations, including specialized query matrices in KQV attention based on attending to either traditional tokens or entities. However, at PT time, LUKE's only modulation over a ROBERTA~\cite{liu_roberta_2019} baseline is an entity masking task.
    \item[\cite{COLAKE_sun-etal-2020-colake}] General domain NLP; COLAKE performs a priori entity linking on the source text, then replaces per-token mentions with entity embeddings, and appends to the input text sub-graphs from a (relational) knowledge graph, including both neighboring mentions and relations in the augmented input text block. This input is then encoded via a transformer that limits attention flow between tokens of different types and trains the entire ensemble with masked language, entity, and relation modelling.
    \item[\cite{PRE_ENC_Xiong2020Pretrained}] General domain NLP; In this paper, traditional masked language modelling is augmented with an entity-replacement-detection task. Named entity recognition and linking are performed before pre-training, and entity replacements are constrained to be the same type as the true entity.
    \item[\cite{LP-BERT_DBLP:journals/corr/abs-2201-04843}] Knowledge Graph Completion; LP-BERT constructs a specialized pre-training corpus consisting of entity-relation statements from a knowledge graph. This is used in a pre-training context under three pre-training tasks: masked language modelling, masked entity modelling, and masked relationship modelling. All three are per-token, and no per-sample tasks are used at pre-training time.
    \item[\cite{UD-PRLM_NEURIPS2021_473803f0}] Multilingual Language Models; UD-PrLM examines multilingual pre-training, and aims to improve it by incorporating universal dependency parse trees into the model. They incorporate a per-token task to align tokens with identified dependency parse tree components, alongside masked language modelling.
\end{description}

\subsection{Language modelling, Per-token KG Integration, \& Supervised Classification}
\begin{description}
    \item[\cite{ERNIE2_sun_ernie_2020,ERNIE3_sun2021ernie}] General domain NLP; ERNIE 2.0 \& 3.0 augments traditional MLM with entity-specific masking (e.g., masking the word ``Mozart'' from the sentence ``Mozart was a musician'') as well as a multi-task per-sample task, largely motivated at classifying a block of text based on internal text cohesion (predict the true order of the sentences within an input sample \& identify whether the sentences within the input sample are spatial neighbors, come from the same document, or come from different documents). ERNIE 3.0 additionally augments pre-training with a per-token relation-embedding task using cloze-filling as a vehicle to perform relation extraction on pre-specified per-token KGs.
    \item[\cite{ERNIE_ALT_zhang_ernie_2019}] General domain NLP; ERNIE (no relation to \cite{ERNIE1_sun_ernie_2019,ERNIE2_sun_ernie_2020}) uses both architectural and objective-function changes to inject per-token knowledge into PT. Specifically, they separately embed all named entities in a sample using the architecture to join contextualized entity embeddings alongside the embeddings of tokens, realizing that entity in the span and performing entity-specific masking. In addition, they simultaneously perform standard MLM and next-sentence prediction in the manner of BERT~\cite{devlin_bert_2019}.
    \item[\cite{MGBERT_behnamghader2021mg}] General domain NLP; MG-BERT introduces a GCNN layer after BERT token, aggregating token embeddings together over a unified graph consisting both of co-occurrence relationships and knowledge graph relationships.
    \item[\cite{JAKET_yu2020jaket}] General domain NLP; JAKET embeds entities by extracting per-token representations of entity texts inside per-entity descriptions, then produces updated KG embeddings via a graph attention network~\cite{GAT_velickovic2018graph}. Those embeddings are then fed into a language model alongside per-token embeddings corresponding to those entities. The entire model is trained according to an MLM objective, plus entity category prediction and relation prediction (only on the entity embeddings extracted from entity descriptions and fed through the GCNN---\emph{not} on the raw entities within the contextualized text).
    \item[\cite{BERTMK_he-etal-2020-bert}] Biomedical NLP; BERT-MK introduces a transformer-based subgraph summarization network that produces entity embeddings for dynamically chosen subgraphs of a given knowledge graph. This network is trained via a contrastive triplet-validity objective. These are then fused with per-token embeddings in free-text based on apriori entity-token matching (\ie, named entity recognition and linking must be performed first and separately before using this model).
    \item[\cite{COKEBERT_su2021cokebert}] General domain NLP; Coke is similar to ERNIE~\cite{ERNIE_ALT_zhang_ernie_2019},  JAKET~\cite{JAKET_yu2020jaket}, and BERT-MK~\cite{BERTMK_he-etal-2020-bert} in that it aggregates entity information by leveraging a GCNN over a restricted dynamic context KG based on token-entity mentions then integrates those augmented embeddings into the per-token embeddings of a BERT-style pretrained model (similar to JAKET and BERT-MK), but also leverages the denoising entity autoencoder task of ERNIE~\cite{ERNIE_ALT_zhang_ernie_2019}. In addition, in the variant of COKE derived from the BERT model, COKE also employs the next-sentence prediction task introduced in BERT~\cite{devlin_bert_2019}.
    \item[\cite{SMEDBERT_zhang2021smedbert}] Medical domain NLP; SMedBERT leverages a complex, multi-faceted loss including MLM, Sentence-order prediction SOP (as introduced in, e.g., ALBERT~\cite{lan_albert_2019}), and includes per-token KG information by aggregating token embeddings across KG embeddings (produced via trans-H~\cite{TRANS_H_wang2014knowledge}) corresponding to matching entities and the neighbors of matching entities in the KG. They also include relation and entity masking variations to ensure the PT model learns per-token information corresponding to the KG. This method bares similarity to Coke~\cite{COKEBERT_su2021cokebert} and JAKET~\cite{JAKET_yu2020jaket}. However, unlike Coke and JAKET, SMedBERT realizes the entity/neighbor matching via a geometric objective, which results in an explicit per-token knowledge graph alignment.
    \item[\cite{Dict-BERT-yu-etal-2022-dict}] General domain NLP; Dict-BERT focuses on augmenting BERT by concatenating definitions of rare words via a per-token KG integration. They add two additional tasks atop the traditional MLM task. First, a task maximizing the mutual information between a masked rare word (treated as a named entity) and its definition (represented as the per-token embedding of the first mention of the entity in the concatenated definition). Second, a task discriminating valid rare word definition per-sequence embeddings from non rare-word definition embeddings via a classification objective.
    \item[\cite{SENTILARE_ke-etal-2020-sentilare}] Sentiment Analysis; SentiLARE integrates sentiment analysis and labels into pre-training by including word polarity signals during masked language modelling and embedding and augmenting pre-training with a supervised sentence sentiment prediction. Word polarities are determined via an external knowledge base integrated at the per-token level.
    \item[\cite{SPIDER_zhang-zhao-2021-structural}] Dialogue Modelling; SPIDER augments traditional MLM and NSP pre-training with two tasks specific to dialogue modelling: first, utterance order prediction, in which individual utterances (which are nested within a larger sample) are shuffled and the true order is predicted, and a geometric task ensuring that subject, verb, object triples from the utterances obey a geometric relationship inspired by KG embedding methods.
\end{description}

\subsection{Language modelling \& Graph link-prediction realized as single-task classification}
These methods all employ some variant of a graph link-prediction task over their data. However, they all realize this link prediction task not by enforcing any relationship between independent sample embeddings but rather by concatenating samples corresponding to linked (or unlinked, for negative samples) pairs of vertices in the source graph, then framing the learning problem as a binary or multi-class classification problem over the (now concatenated) single output whole sample embedding. In doing so, they transform the task from one that implies a deep geometric constraint over the output latent space to one that only enforces an intra-sample objective and imposes only a shallow geometric constraint on the per-sample latent space.

\begin{description}
    \item[\cite{devlin_bert_2019}] General domain NLP; Masked language model plus the binary classification of whether the input text block is sequentially consistent, with samples chosen via true positive pairs vs. randomly joined sentences. This can be seen as a link prediction task over a graph consisting of independent, disconnected ``sticks'', with each stick corresponding to sentences in the documents in the corpus, in sequential order.
    \item[\cite{lan_albert_2019}] General domain NLP; Masked language model plus the binary classification of whether the input text block is sequentially consistent, with samples chosen via true positive pairs vs. reordered positive sentence pairs. This can be seen as a link prediction task over a directed graph consisting of independent, disconnected ``sticks'', with each stick corresponding to sentences in the documents in the corpus, in sequential order, with edge direction indicating sequential ordering.
    \item[\cite{LINKBERT_yasunaga-etal-2022-linkbert}] General domain NLP; Masked language model plus the classification of whether the input text block contains sentences from either (1) random documents, (2) a sequentially consistent pair within a single document, or (3) within a pair of sentences within two linked documents according to a document linking graph $G$. This can be seen as a link prediction/edge classification task over a graph whose nodes are text blocks in the corpus, with two distinct edge modalities. First, to capture sequential consistency within a document, one edge type produces a set of independent, disconnected ``sticks'', with each stick corresponding to sentences in the documents in the corpus, in sequential order. Second, to capture the document linking graph $G$, sentences in a document $D_i$ are all linked to all sentences in a document $D_j$ if and only if documents $i$ and $j$ are linked in $G$.
    \item[\cite{kuncoro-etal-2020-syntactic}] General domain NLP; While this model incorporates an interesting per-token syntatic knowledge distillation procedure, at a per-token level it merely leverages BERT's NSP loss~\cite{devlin_bert_2019}.
\end{description}

\subsection{Language modelling \& Single-task Classification}
\begin{description}
    \item[\cite{PLUS_min_pre-training_2020}] Protein sequences; Masked language model plus the multi-class classification of to which protein family an input sequence belongs. Uses non-standard whole-sequence embedding procedure (no \texttt{[CLS]} token).
    \item[\cite{STRUCTBERT_Wang2020StructBERT}] General domain NLP; StructBERT includes masked language modelling, a token permutation language modelling task, and an extended version of the NSP/SOP task at a per-sample level.
\end{description}

\subsection{Language modelling \& Multi-task Classification}
\begin{description}
    \item[\cite{MT_DNN_liu_multi-task_2019}] General domain NLP; Masked language model plus multi-task classification across various NLP tasks.
    \item[\cite{GRAPH_PT_hu_strategies_2019}] Graph data; This model uses a masked imputation task similar to a masked language model and a highly multi-task supervised whole-graph level prediction. On this non-NLP domain, \cite{GRAPH_PT_hu_strategies_2019} finds that the multi-task whole-graph level task is essential for performance.
    \item[\cite{mcdermott_comprehensive_2020}] EHR Timeseries data; This model uses a masked imputation task similar to a masked language model over time series data and a multi-task supervised whole-sequence prediction task. On this non-NLP domain, \cite{mcdermott_comprehensive_2020} finds the multi-task whole-sequence level task essential for performance.
\end{description}

\subsection{Language modelling \& whole-sample graph-based contrastive objectives}
\begin{description}
    \item[\cite{wang_kepler_2019}] General domain NLP; KEPLER augments traditional MLM on text samples with a constraint ensuring the (per-sample) embeddings of entity descriptions pulled from pre-specified knowledge graphs (KGs) reflect geometric constraints, leveraging the \cite{sun2018rotate} geometric constraints. As we will see in our theoretical analyses, these constraints are much more restrictive on the latent space geometry and thus imply a greater encoding of domain knowledge in the model. Note that JAKET~\cite{JAKET_yu2020jaket} also leverages entity descriptions in its per-token encoding. However, these descriptions are (1) extracted via per-token embeddings, using the first mention of the token, not whole-sample embeddings, and (2) integrated back into the original text in a per-token manner, not optimized over directly via geometric constraints as in KEPLER.
    \item[\cite{CK_GNN_fang2021knowledge}] Molecules; CK-GNN designs a pre-training scheme for molecular graphs in which a molecular GNN is trained to produce molecule embeddings that obey the similarity structure of a 1-NN graph in a cluster-limited molecular fingerprint space (using the Dice similarity coefficient). Unlike the NLP approaches, this method has no intra-sample (\ie, per-token, where here ``token'' refers to individual atoms within the molecular graph) pre-training task.
    \item[\cite{XLM_K_jiang2022xlm}] Multi-lingual NLP; Much like KEPLER, XLM-K augments traditional MLM with two tasks that constrain the geometry of the per-sample latent space via a (now multi-lingual) graph of entity descriptions linked to sentences containing said entities. Like KEPLER, as the graph connections here are defined only for entity descriptions and not all free-text, the latent space regularization is only over a limited slice of the space.
    \item[\cite{WEBFORMER_guo2022webformer}] General domain NLP/IR; WebFormer designs a pre-training scheme leveraging the structure of DOM trees in HTML pages to impose multiple per-sample and per-sample/per-token hybrid constraints that encourage individual samples to be (a) close to noised versions of themselves based on reordering or masking and (b) to be close to representations of their parent/child nodes in the DOM tree, thus imposing a structural penalty geometrically. By mixing per-sample and per-token tasks, WebFormer even more closely entangles the per-sample and per-token latent spaces in their model, and this approach bears closer study in other contexts.
\end{description}

\subsection{Language modelling \& whole-sample augmentation/noising based contrastive objectives}
\begin{description}
    \item[\cite{INFOWORD_Kong2020A}] General domain NLP; InfoWord incorporates an objective alongside masked language modelling which pushes the whole-sample embedding of a sentence to have high mutual information with various sub-contexts within that sentence and low mutual information with sub-contexts of other sentences.
    \item[\cite{DECLUTR_giorgi-etal-2021-declutr}] General domain NLP; DeCLUTR optimizes for masked language modelling alongside a contrastive objective comparing anchor spans to positive spans chosen from within individual samples, contrasted against spans from other samples. This is considered ``whole-sample'' rather than a per-token contrastive loss as the embeddings of the spans (which can be quite long) are produced via a canonicalized pooling operation used for sentence embeddings.
    \item[\cite{CLEAR_wu2020clear}] General domain NLP; CLEAR optimizes for masked language modelling alongside a contrastive objective powered by per-sentence noising strategies, including word or span deletion, reordering, and synonym substitution.
    \item[\cite{COCO_LM_NEURIPS2021_c2c2a045}] General domain NLP; COCO-LM builds on other discriminative language modelling variants such as ELECTRA~\cite{ELECTRA_clark_electra_2019} by adding two additional tasks. First, a true language modelling task atop the auxiliary-model-driven corrupted input text. Second, a contrastive objective pushing corrupted sentences towards their un-corrupted originals and those derived from distinct sentences farther apart.
    \item[\cite{STS_CT_carlsson2021semantic}] General domain NLP; Semantic re-tuning via contrastive tension adds a pre-training objective onto language model pre-training. This is done to encourage the final per-sample representations of a single sentence embedded via two otherwise independently trained models to be similar and those of different sentences to be distinct.
    \item[\cite{KCL_fang2022molecular,GraphCL_NEURIPS2020_3fe23034,JOAO_pmlr-v139-you21a,GCC_10.1145/3394486.3403168,MICRO_GRAPH_zhang2020motif}] Networks; KCL,GraphCL, JOAO, MICRO-Graph and GCC use augmentation-based contrastive learning pre-training methods for network datasets. KCL is notable as it is (1) specialized for molecular graphs and (2) uses a knowledge-derived augmentation strategy that constructs a knowledge enriched version of an input molecular graph as its ``augmentation policy.'' MICRO-Graph is also notable as its contrastive objective compares a graph to dynamically clustered ``motif'' subgraphs from within said graph as positive pairs.
    \item[\cite{GLM_shen-etal-2020-exploiting}] General domain NLP; GLM integrates a per-token KG through traditional entity masking (albeit with an improved selection mechanism) and a per-sample contrastive objective that uses the entity knowledge graph to generate distractor negative samples for the contrastive learning task.
    \item[\cite{CAPT_luo2020capt}] General domain NLP \& Computer Vision; CAPT proposes a noising based contrastive learning loss \emph{in substitution for} the masked language modelling loss of BERT. They employ no per-token pre-training task.
    \item[\cite{GEARNET_zhang2022protein}] Protein Sequences/Structures; GearNet introduces a vehicle for pre-training not over protein sequences, but rather over protein structures, realized as graphs. They combine intra-sample/per-amino-acid tasks, including prediction of masked node features and prediction of geometric relationships between nodes as implied by the protein graphs, and a per-sample noising based contrastive objective.
\end{description}

\subsection{Language modelling \& multi-modal or multi-lingual contrastive objectives}
Note that by viewing multiple data modalities as ``augmentations'' of the data samples, one can realize these methods (in general) as examples of augmentation-based contrastive learning objectives, such as those used in \cite{SIMCSE_gao-etal-2021-simcse}. However, as these methods are common, we highlight them explicitly here.
\begin{description}
    \item[\cite{INFOXLM_chi-etal-2021-infoxlm}] General domain NLP; InfoXLM focuses on multi-lingual pre-training, and leverages per-token tasks. This includes multi-lingual masked language modelling and translation language modelling (\ie, variations on a traditional masked language modelling task). It also incorporates a cross-lingual per-sample contrastive objective that aligns the geometry of the latent spaces across distinct languages. One important nuance is that they use different layer depths to define the latent space for their cross-lingual contrastive objective vs. their per-token objectives, which is not natively describable in our framework. In addition, as each monolingual corpus lacks any rich, independent per-sample task, any individual monolingual latent space cannot be guaranteed to have any rich structural constraints. 
\end{description}

\subsection{Language modelling alone with relationally-concatenated samples}
These methods concatenate samples together before processing them with a pre-training encoder based on inter-sample relations. This is an orthogonal direction to adding greater per-sample dependencies to pre-training methods than our framework but warrants commentary nonetheless.

\begin{description}
    \item[\cite{MSA_pmlr-v139-rao21a}] Protein sequences; MSA transformers extend protein-sequence language models such that they do not take in as input a single sequence but rather an entire multiple-sequence alignment (MSA) profile. These profiles consist of many sequences corresponding to evolutionary homologs of the same protein. This concatenated input is processed via a sparsified form of axial self-attention, which enables cross-attention between the various aligned sequences. They impose no per-sequence tasks by default in this architecture.
    \item[\cite{KNN_PT_levine2022the}] General domain NLP; This theoretical analysis shows that transformers cannot model dependencies between sentences that never appear in the same example during pre-training. To combat this, they propose concatenating samples via inter-sample relations (in particular, via a kNN method) at pre-training time, enabling a greater diversity of cross-attention contexts during pre-training vs. fine-tuning. Thus, while they only use language modelling during pre-training, they speculate that their sample-augmentation procedure helps the model better reason about per-sample information through per-token tasks.
    \item[\cite{CDLM_caciularu-etal-2021-cdlm-cross}] General domain NLP; CDLM proposes to concatenate multiple related documents (leveraging categorical information to cluster documents) together into a single sample prior to performing traditional masked language modelling. To limit the model's complexity, attention is restricted to intra-document for unmasked tokens but allowed to be global for masked tokens.
    \item[\cite{REALM_10.5555/3524938.3525306}] General domain NLP; REALM uses a latent variable model to learn a relevance score between input text spans and documents in an auxiliary document base. The top-$k$ documents, according to this relevance score, are then concatenated to the input prior to solving the masked language modelling task used during pre-training. In this way, the model learns to join relevant documents from an external knowledge base in accordance with which documents would most improve the masked language modelling objective. In addition, by learning this relevance score, the model introduces an implicit whole-sample structural constraint on the latent space according to the unsupervised clustering induced by relevance assignment. 
\end{description}

\subsection{Autoencoding \& Unsupervised Clustering}
\begin{description}
    \item[\cite{MARGE_lewis2020pre}] General domain NLP; MARGE deviates significantly from the norm by not employing any form of language modelling or other forms of a per-token pre-training task. Instead, it employs only a per-sample contextualized autoencoding objective and an unsupervised per-sample retrieval step (to provide context for said autoencoding). While this approach does provide a deeper form of a per-sample structural constraint than many other approaches, it is also implicit and has no mechanism for injecting domain knowledge. MARGE is also tested solely on downstream tasks at the per-sample level, so it is unclear if this method would offer reduced benefits for per-token downstream tasks.
\end{description}

\subsection{Methods orthogonal to our framework}
\begin{description}
    \item[\cite{KGBART_liu2021kg}] KG-BART is a text-generation model that leverages per-token knowledge after a text-encoder to enrich the generated text with information from a textual knowledge graph (in a per-token manner). It is neither used for general pre-training nor does it leverage any additional per-sample constraints.
    \item[\cite{GRAPHFORMERS_yang2021graphformers}] Text-based Knowledge Graphs; This work produces embeddings of nodes in KGs by combining transformer-based text encodings with graph convolutional network KG embedding methods, leveraging link prediction as the pre-training task. Entity descriptions / textual features represent the individual nodes. Link prediction can be seen as inducing a geometric constraint via the connectivity of the knowledge graph on whole-sample embeddings. However, given that relationships are used in encoding the data as well, GraphFormer cannot be used in a context where KG links may not be observed at FT time. It should be seen not as a general text PT method but as an advanced KG embedding mechanism, so it does not directly fall under our framework.
    \item[\cite{KeLM_agarwal-etal-2021-knowledge}] KeLM (unrelated to KELM~\cite{KELM_lu2021kelm}) is a method for converting a free-text KG into textual nodes so language modelling can be used over that corpus and is orthogonal to the methods of pre-training.
    \item[\cite{AKEBERT_ribeiro2021combining}] This paper is a method for populating a KG from free-text via BERT. It has no bearing on incorporating structure or knowledge into PT and is irrelevant to our framework.
    \item[\cite{SISRKG_zhang2021drop}] This paper presents a method to drop redundant triples from a knowledge graph and a regularization technique to limit the impact of added irrelevant knowledge to per-token knowledge-enhanced PT methods such as ERNIE~\cite{ERNIE_ALT_zhang_ernie_2019}.
    \item[\cite{KGBERT_yao2019kg}] Knowledge Graph Completion; KG-BERT is a method for knowledge graph completion in which textual representations of entities and relations in KGs are embedded by fine-tuning a pre-trained BERT style transformer for link prediction over a given KG. As this is only for knowledge graph completion, it is orthogonal to our study of pre-trained models in general.
    \item[\cite{SimKGC_wang-etal-2022-simkgc}] Knowledge Graph Completion; Much like KG-BERT, SimKGC is a method for knowledge graph completion that fine-tunes a BERT model via a contrastive loss over a fixed knowledge graph for link prediction. Though their methodology overlaps with ours in that both use variants of contrastive losses and SimKGC explores more complex negative sampling strategies, the two methods are still very different. Ours is focused on general pre-training and uses a single encoder and a unified latent space. In contrast,  SimKGC is only examined for KG completion and encodes head and tail entities via separate encoders.
    \item[\cite{CLEVE_wang-etal-2021-cleve}] Event Extraction (EE); CLEVE designs a pre-training method specifically for event extraction. Their pre-training method includes a text-encoder which includes a \emph{cross-event} contrastive loss pushing \emph{individual tokens} from the same ``event'' closer together than those from different events, which bears a surface similarity to our approach. In addition, they add a graph encoder over the semantic structure of events. Their methodology is focused solely on EE, which is orthogonal to our more general PT framework.
    \item[\cite{VILT_pmlr-v139-kim21k}] General domain NLP and Computer Vision; ViLT is a method for pre-training aligned text-image pairs. It leverages masked language modelling, an image-text matching binary classification objective, and a contrastive objective comparing image and text representations. This multi-modal contrastive objective is very similar (insofar as it relates to our framework) to those works that perform multi-lingual or other multi-modal contrastive methods. In ViLT, however, the transformer architecture processes images and text jointly in a single encoder, so it is not well suited for use on only images or only text. This, combined with its focus on computer vision, renders it orthogonal to our framework.
    \item[\cite{STRUCTURALLM_li-etal-2021-structurallm}] General domain NLP and Computer Vision; StructuralLM proposes a new method of pre-training for scanned documents that takes advantage of the structure of the document w.r.t. images and text simultaneously. As their focus is on cross-modal pre-training of text and image alignment, it is orthogonal to our work.
    \item[\cite{Cont-Disc-PT_fan2022unified}] General domain NLP and Computer Vision; This paper proposes a framework for simultaneous (and continuous) discovery of edges in a multi-modal knowledge graph and the leveraging of that knowledge graph to inform representation learning. However, it is not suitable for our framework for two reasons. First, like ViLT, it is focused on image-text alignment pre-training. Second, when producing node (\eg, images or text snippets) representations, it requires connectivity information in the associated multi-modal knowledge graph. In contrast, our methods take as input only elements from $\mathcal X$.
    \item[\cite{SAPBERT_liu-etal-2021-self}] Named Entity Linking; SapBERT is a method for aligning the output of a pre-trained language model with a per-token knowledge graph through a metric learning loss applied at a per-sample level but only over entity names (not even entity descriptions). As it applies this as a secondary, post-PT stage, and this method only optimizes for alignment between entity names and a static KG, it is not a general PT framework. It is thus orthogonal to our efforts here.
    \item[\cite{HARP_10.1145/3459637.3482286}] Information Retrieval; HARP is a method for specializing pre-training towards ad-hoc query information retrieval. They introduce four retrieval-specific pre-training tasks leveraging hyperlinks in Wikipedia articles in addition to traditional masked language modelling. Rather than using the raw text of the hyperlinks or the per-sample representations of text spans containing hyperlinks, both of which are explored in~\cite{WIKILINK_XLNG_calixto-etal-2021-wikipedia}, these authors use attention weights to extract various ``queries'' from the underlying text and match those against possible destination pages via contrastive losses. This, therefore, does not impose a constraint on the latent space over the original pre-training dataset $\mathcal X$ (but instead introduces a new latent space consisting of query spans) and is further specialized exclusively for ad-hoc retrieval tasks.
    \item[\cite{CPT_HG_10.1145/3459637.3482332}] Node Embedding for Heterogeneous Graphs; CPT-HG is a contrastive pre-training framework to embed nodes in a heterogeneous network. Unlike in our setting, where the pre-training graph $G_\PT$ is \emph{only used as an implicit input to derive the loss function}, in CPT-HG the graph (with entire edge connectivity information) \emph{is} the input to the problem. Thus, node embeddings will rely on connectivity information, which is not permissible in our pre-training context. So, this method is orthogonal to our study here.
    \item[\cite{CODE_chen2022code}] Expert Matching; CODE is a method specifically and exclusively designed to discover appropriate experts in an employment/contracting setting and is thus orthogonal to our framework, which is focused on more general pre-training.
\end{description}

\subsection{Methods that only change things at FT time}
\begin{description}
    \item[\cite{MOP_meng2021mixture}] Biomedical domain NLP; MOP does not change anything at PT time but trains sub-KG adapters on entity recognition tasks prior to FT to infuse entity knowledge into the PT method. It is a per-token pre-training method. 
    \item[\cite{KBERT_liu2020k}] General domain NLP; K-BERT, at PT time, is actually equivalent to BERT~\cite{devlin_bert_2019}. However, it does do other interesting things at FT time, including augmenting the sentence flow with injected per-token knowledge graphs and limiting self-attention to only flow along links supported by the original sentence or the injected knowledge. However, as this is only true at FT time, it is equivalent to BERT at PT time. 
    \item[\cite{SBERT_reimers-gurevych-2019-sentence}] General domain NLP; This model, at PT time, is equivalent to BERT~\cite{devlin_bert_2019}. Like \cite{KBERT_liu2020k}. However, it specializes in a fine-tuning procedure for sentence information retrieval tasks, similar to how PT is adapted in this framework.
    \item[\cite{CONSERT_yan-etal-2021-consert}] General domain NLP; ConSERT adds an auxiliary specialization stage after pre-training to fine-tune sentence representations. This new stage imposes a SimCLR~\cite{SIMCLR_chen_simple_2020} style data-augmentation/noise-invariance based contrastive learning objective, using adversarial perturbations, token shuffling, token/feature/span erasure, and dropout noising methods.
    \item[\cite{IS_BERT_zhang-etal-2020-unsupervised}] General domain NLP; IS-BERT does not modify anything from traditional BERT at pre-training time. However, they add a second PT stage to optimize sentence representations alone using an auxiliary feature extractor in the form of various CNNs applied atop BERT token representations. The final sentence representation is trained to maximize mutual information with various sub-contexts within the sentence but low mutual information with other sentences. In this second pre-training stage, there is no language modelling performed. As this approach only adapts an auxiliary featurizer to produce sentence encodings and is not intended for general transfer learning, it is inappropriate for our framework. A similar work that integrates both components during pre-training, and thus is relevant in our work is \cite{INFOWORD_Kong2020A} and is discussed above.
    \item[\cite{KELM_lu2021kelm}] General domain NLP; KELM does not modify PT objective but instead enhances a model at FT time by injecting per-token knowledge via a GNN module atop the pre-trained LM embeddings via a unified text-entity graph. It is similar to KBERT~\cite{KBERT_liu2020k} in this way but resolves other issues with that approach relating to knowledge ambiguity and by supporting multi-hop reasoning, again over the per-token embeddings.
    \item[\cite{KIBERT_faldu2021ki}] General domain NLP; KI-BERT augments BERT with KG-specific information via joint token-entity embeddings and information fusion but does this only at FT time.
    \item[\cite{KXLNET_yan2021general}] General domain NLP; K-XLNet introduces a secondary FT stage in which knowledge injectors throughout an XL-Net architecture are further trained to leverage knowledge (encoded via free-text entity descriptions) that is injected into input sentences alongside matched tokens. It does not modify the XL-Net PT stage at all.
    \item[\cite{KADAPTER_wang2020k}] General domain NLP; K-Adapter proposes to pre-train various knowledge adapters that can be used alongside a pre-trained language model at a fine-tuning time. Thus, while there is a pre-training process for the adapters, this process does not modulate the original pre-trained language model. In addition, both adapters pre-trained in this work are based on per-token knowledge graphs; one leverages concatenated entity embeddings to perform relation classification, and another predicts which token in the sentence is the ``head'' in a dependency parse tree, so no per-sample constraints are applied.
    \item[\cite{EBERT_poerner-etal-2020-e}] General domain NLP; E-BERT injects per-token knowledge into BERT by first aligning embeddings of a knowledge graph with the input word piece embedding space of a (fixed, pre-trained) BERT model, then using various strategies to input them alongside their source mentions in FT text. They do no additional pre-training, so this model only affects the model at FT time.
    \item[\cite{gunel2021supervised}] General domain NLP; \cite{gunel2021supervised} augment LMPT methods with an additional, pre-FT procedure in which the model is further trained using a supervised, per-sample metric learning task leveraging FT labels directly to form the classes used for metric learning. They do not materially change the task-independent PT procedure, though their FT metric learning procedure does induce some structure at the per-sample level.
    \item[\cite{GREASELM_zhang2022greaselm}] QA; GreaseLM is a method for fusing information from knowledge graphs into pre-trained language models. It shares many similarities with methods that do this for pre-training purposes, such as JAKET~\cite{JAKET_yu2020jaket}, CokeBERT~\cite{COKEBERT_su2021cokebert}, SMedBERT~\cite{SMEDBERT_zhang2021smedbert}, and Bert-MK~\cite{BERTMK_he-etal-2020-bert}. However, unlike these methods, it only employs these techniques at the fine-tuning time, for question answering tasks specifically. As it is not focused on general pre-training, it is outside our scope.
    \item[\cite{KNN_LMS_Khandelwal2020Generalization}] Language modelling; kNN language models improve the text generation powers of language models by augmenting traditional decoding with a nearest-neighbor lookup operation over a text datastore leveraging the embeddings of a token's leftward context by the language model to judge nearest neighbors. However, it involves no additional language model training and can only be applied at the fine-tuning time to aid in text generation, and is thus out of our scope.
    \item[\cite{NT_XENT_kim-etal-2021-self}] Sentence embedding; NT-Xent proposes a secondary specialization stage after pre-training only for generating sentence embeddings. To do this, they employ a contrastive objective contrasting the final CLS embeddings of an updating, specialized BERT model against a pooled aggregate of the per-token embeddings across all layers of the pre-trained BERT model used to initialize the specialized sentence embedding model.
    \item[\cite{BERT_FLOW_li-etal-2020-sentence,BERT_WHITENING_su2021whitening,WHITENING_BERT_huang-etal-2021-whiteningbert-easy}] Sentence Embedding; These methods propose to use unsupervised per-sample smoothing operations (a normalizing flow network in \cite{BERT_FLOW_li-etal-2020-sentence} and a mean/covariance standardization whitening operation in \cite{BERT_WHITENING_su2021whitening,WHITENING_BERT_huang-etal-2021-whiteningbert-easy}) on the per-sample embeddings after pre-training in order to produce higher quality per-sample embeddings.
    \item[\cite{SIMCSE_gao-etal-2021-simcse}] General domain NLP; SimCSE extends traditional MLM by imposing a second pre-training stage for optimizing sentence embeddings. In this stage, SimCSE optimizes the transformer such that the whole-sample embeddings satisfy either a supervised or unsupervised contrastive learning objective. In the supervised case, this is based on labeled sentence pairs according to a Natural Language Inference (NLI) task, with entailment pairs being treated as positives and contradiction pairs as hard negatives. In the unsupervised case, this is based solely on applying multiple dropout masks to the same sentence to generate positive pairs. Any two distinct sentence inputs are treated as negative samples. This extra pre-training stage is applied to a relatively small number of samples ($10^6$) relative to the entire PT cohort, which may help prevent catastrophic forgetting of the original pre-training objective.
    \item[\cite{SPECTER_cohan-etal-2020-specter}] Academic NLP; SPECTER extends traditional language model pre-training by imposing a second pre-training stage for optimizing document embeddings (realized as \texttt{[CLS]} token embeddings of concatenated academic paper titles and abstracts). This stage uses a triplet-based geometric loss to ensure that these per-sample embeddings reflect the structure of a pre-specified citation network. This is a form of an explicit, structural constraint; however, they do not ever test fully fine-tuning the SPECTER model in their paper and only compare it against other, frozen pre-trained language models. This is likely to have a significant impact on model comparisons. Similar to SimCSE~\cite{SIMCSE_gao-etal-2021-simcse}, this extra pre-training stage is applied to a small number of samples (146K documents) to help prevent catastrophic forgetting of the original pre-training objective.
    \item[\cite{WIKILINK_XLNG_calixto-etal-2021-wikipedia}] General domain NLP; This paper introduces a second pre-training stage after multi-lingual masked language modelling. In this second stage, hyperlinks in the source text (drawn from Wikipedia) are matched via single-task classification to a curated set of destination URL categories, collapsing all URLs pointing to the same Wikipedia page across languages into one. They do this classification in several ways, including incorporating the per-sample representation of the text span rather than merely the hyperlink token representations themselves (likely motivated by the likelihood of only a single hyperlink being present in the source text). We can realize this task as instances of several other common paradigms: (1) Single-task classification applied to the per-sample representation, (2) link prediction in a graph linking cross-lingual Wikipedia pages together, or (3) as an example of named entity recognition. This second stage is only allowed to modify the last two layers of the transformer architecture, which may be a vehicle to prevent catastrophic forgetting.
    \item[\cite{SAKG_BERT_9490220}] Sentiment Analysis; SAKG-BERT augments a pre-trained language model with a sentiment-analysis knowledge graph at the fine-tuning time only by concatenating relevant relationships from the KG based on sentiment-laden terms appearing in the review to the raw input text. They do not otherwise change the pre-training or fine-tuning process.
\end{description}